\documentclass{article}

\usepackage[preprint]{neurips_2023}

\usepackage{iftex}
\ifPDFTeX
  \usepackage[T1]{fontenc}
  \usepackage[utf8]{inputenc}
  \usepackage[varqu]{zi4}
\else
  \usepackage{fontspec}
\fi

\usepackage{amsmath,amssymb,amsthm}
\usepackage{graphicx}
\usepackage{booktabs}
\usepackage{multirow}
\usepackage{xcolor}
\usepackage{tikz}
\usetikzlibrary{arrows.meta,positioning,fit,backgrounds,calc,shapes.geometric}
\usepackage{algorithm}
\usepackage{algpseudocode}
\usepackage{listings}
\lstdefinestyle{prompt}{basicstyle=\footnotesize\ttfamily,frame=single,rulecolor=\color{black!30},backgroundcolor=\color{black!4},breaklines=true,columns=fullflexible,keepspaces=true,xleftmargin=5pt,xrightmargin=5pt,framerule=0.4pt,aboveskip=6pt,belowskip=4pt}
\usepackage[colorlinks=true,linkcolor=blue,citecolor=blue,urlcolor=blue]{hyperref}

\newcommand{\rc}{\textsc{LAFP}}                       

\title{LLM as Forecasting Planner: Training-Free Text Conditioning for Time-Series Foundation Models}

\author{%
  Huu Hiep Nguyen \quad Dung Nguyen \quad Minh Hoang Nguyen \quad Dai Do \quad Hung Le \\
  Applied Artificial Intelligence Initiative \\
Deakin University \\
Geelong, Australia \\
}

\begin{document}

\maketitle

\begin{abstract}
Text-conditioned time-series forecasting predicts a series from both its numerical history and natural-language context, allowing forecasts to account for events and constraints that the past alone cannot reveal.
This requires both reliable numerical forecasting and the ability to interpret contextual information.
Time-series foundation models (TSFMs) provide strong numerical forecasts, while large language models (LLMs) can reason over text, but combining their strengths remains challenging because asking an LLM to generate or revise forecast values directly can distort the temporal structure captured by the TSFM.
We instead formulate forecasting as a planning problem over TSFM-generated trajectories.
The frozen TSFM acts as a simulator that proposes numerical continuations, while the LLM acts as a policy and value function that guides candidate selection and evaluates completed trajectories against the context.
We instantiate this as \rc{} (\textbf{L}LM \textbf{A}s \textbf{F}orecasting \textbf{P}lanner), a training-free framework that bridges the modality gap without retraining either model, using Monte Carlo tree search (MCTS) over the forecast
horizon with a \emph{Ranker} LLM as policy and a \emph{Judge} LLM as value function.
Experiments on Context-is-Key and Time-MMD across two TSFM backbones (Chronos and TimesFM) and four LLMs show that \rc{} delivers consistent improvements across model choices, supporting sequential search as an effective training-free approach to text-conditioned forecasting.
\end{abstract}


\section{Introduction}
\label{sec:intro}

Time-series forecasting supports decision-making across domains including energy, public health, finance, and supply chains. Decades of statistical and deep learning research have improved the ability to predict a series from its numerical history~\citep{nie2023patchtst, ansari2024chronos, woo2024moirai, das2024timesfm}. However, numerical history alone cannot reveal many events and constraints that shape the future, such as plant shutdowns, policy changes, or heat-wave advisories, even when this information is available in text. Human analysts rarely forecast from numbers in isolation, instead interpreting them alongside news, reports, and domain knowledge. This motivates \emph{text-conditioned time-series forecasting}, which predicts a series from both its numerical history and accompanying natural-language context, allowing relevant textual information to refine forecasts beyond what the past alone can support~\citep{williams2024cik, liu2024timemmd, ashok2025beyond}.

Text-conditioned time-series forecasting requires both reliable numerical forecasting and contextual reasoning. TSFMs provide the former. Pretrained on large and diverse time-series corpora, models such as Chronos~\citep{ansari2024chronos}, Moirai~\citep{woo2024moirai}, and TimesFM~\citep{das2024timesfm} can generate plausible futures while preserving temporal properties such as scale, seasonality, and uncertainty. However, most TSFMs condition only on numerical history and cannot directly incorporate textual information about future events or external constraints. LLMs provide the complementary capability of interpreting such information and reasoning about its implications for possible futures.

\paragraph{LLM-as-forecaster.}
Existing training-free approaches commonly exploit this capability by casting the LLM as a forecaster and asking it to generate or revise numerical predictions directly~\citep{gruver2023llmtime}, sometimes through chain-of-thought reasoning or multi-agent collaboration~\citep{das2026nexus, ashok2025beyond}. This role assignment is problematic because LLMs are not reliable numerical time-series models. Language pretraining does not equip them to capture temporal dependencies consistently~\citep{tan2024useful}, and controlled evaluations show that their time-series reasoning performance is only slightly above chance~\citep{merrill2024}. The central question is therefore not whether LLMs can contribute, but how their contextual reasoning should be used. Existing methods give them direct control over forecast values rather than restricting them to the interpretation of textual context.


\paragraph{LLM as forecasting planner.}
We propose to separate contextual reasoning from numerical generation. The TSFM remains responsible for proposing plausible numerical futures, while the LLM determines which of these futures are most consistent with the textual context. We formulate this separation as a planning problem in which a simulator generates candidate states, a policy guides the choice of actions, and a value function evaluates completed trajectories~\citep{lightman2023verify}. In our setting, the TSFM serves as the simulator by generating candidate numerical continuations. The LLM provides the policy by ranking these continuations at each step and the value function by evaluating completed trajectories against the context. Numerical generation therefore remains entirely with the TSFM, while the LLM influences the forecast only through selection and evaluation. This division of roles requires no additional training and can be applied across different TSFM and LLM backbones.


We instantiate this framework as \rc{} (LLM As Forecasting Planner). Rather than reranking a fixed pool of complete forecasts, \rc{} applies MCTS over a sequence of forecast windows, with several TSFM-generated continuations considered at each step. This allows promising prefixes from different TSFM samples to be explored and combined. The LLM has two roles. The \emph{Ranker} ranks candidate continuations during search and provides the policy prior. The \emph{Judge} evaluates complete trajectories after rollout and provides the value signal. Each simulation consists of five stages: \textbf{Selection}, where PUCT chooses a path using the Ranker prior; \textbf{Expansion} generates new continuations with the TSFM; \textbf{Ranking}, where the Ranker ranks these candidates; \textbf{Simulation}, where the TSFM completes the selected trajectory; and \textbf{Backpropagation}, where the Judge scores the completed trajectory and the score is propagated through the tree. After a fixed number of simulations, the final forecast is obtained from the accumulated search state.


In summary, our contributions are:
\begin{itemize}
\item We formulate text-conditioned time-series forecasting as a planning problem over TSFM-generated trajectories. The TSFM acts as the numerical simulator, while the LLM provides policy and value signals for selecting and evaluating candidate futures.
\item We propose \rc{}, a training-free MCTS framework that combines a frozen TSFM with a frozen LLM. A \emph{Ranker} guides the search over candidate continuations, a \emph{Judge} evaluates completed trajectories, and MCTS reuses these evaluations to retain and extend promising forecast prefixes. The framework is agnostic to the choice of TSFM and LLM backbone.
\item We provide a comprehensive evaluation of \rc{} on Context-is-Key and Time-MMD across two TSFM backbones and four Ranker/Judge LLMs. The evaluation includes comparisons with context-blind TSFMs, direct LLM forecasters, and Best-of-$N$ reranking, together with component ablations and analyses of search budget, robustness, and Judge behaviour. The results show that \rc{} consistently improves on the corresponding context-blind TSFM and generally outperforms flat reranking across model configurations.
\end{itemize}

\section{Related Work}
\label{sec:related}

\paragraph{Time-series foundation models.}
Time-series foundation models (TSFMs) are pretrained on large and diverse time-series datasets for zero-shot forecasting across domains~\citep{ansari2024chronos, woo2024moirai, das2024timesfm, rasul2023lagllama, garza2023timegpt}. They use different architectures and representations. Chronos~\citep{ansari2024chronos} discretises scaled values and trains a T5-based autoregressive model. Moirai~\citep{woo2024moirai} and Moirai-MoE~\citep{woo2024moiraimoe} support forecasting across different numbers of variables and sampling frequencies. TimesFM~\citep{das2024timesfm} uses a decoder-only architecture with patch-based inputs and quantile outputs. These models learn temporal patterns from numerical histories but generally do not incorporate textual context. \rc{} builds on their complementary strength in numerical forecasting by using a frozen TSFM as the simulator that generates candidate continuations, while delegating context-dependent selection and evaluation to the LLM. This formulation is independent of the particular TSFM architecture or representation.

\paragraph{Text-conditioned time-series forecasting.}
Text-conditioned forecasting requires a model to use both numerical history and natural-language context. Context-is-Key (CiK)~\citep{williams2024cik} contains 71 tasks designed so that the context provides information not available in the numerical history. Performance is measured using region-of-interest CRPS. Time-MMD~\citep{liu2024timemmd} pairs time series from nine domains with textual reports, and TaTS~\citep{li2026language} provides a standardised evaluation protocol for this benchmark. TimesX~\citep{liu2026timesx} argues that current context-enriched benchmarks remain limited in scale and in the types of context they cover, and introduces a larger, automatically constructed alternative. Existing methods either train modules to fuse the two modalities~\citep{liu2024timemmd, nguyen2026spectral, nguyen2026doestextactuallyhelp} or prompt an LLM to generate or revise forecast values~\citep{gruver2023llmtime, ashok2025beyond, das2026nexus}. The former requires additional multimodal training, whereas the latter places numerical generation in the hands of the LLM. Closest to our setting, the concurrent PostTime~\citep{liu2026posttime} post-trains an LLM with supervised fine-tuning and reinforcement learning to act as a context-guided revisor that decides whether to revise or preserve the forecast of a TSFM. This keeps a numerical backbone in the loop, but it requires a dedicated post-training stage, and the revised forecast values are still generated by the LLM. This leaves an unresolved gap: how to incorporate textual context without any training while retaining the numerical forecasting capabilities of a pretrained TSFM and keeping forecast values under TSFM control. \rc{} addresses this gap by using a frozen TSFM to generate candidate forecasts and an LLM to guide sequential search over them.


\paragraph{LLMs for time-series forecasting.}
Many training-free methods use an LLM as a forecaster and ask it to generate numerical predictions from the time-series history and associated context. LLMTime~\citep{gruver2023llmtime} represents a time series as a sequence of numerical tokens and generates future values autoregressively. NEXUS~\citep{das2026nexus} combines predictions from multiple LLM agents, while other work studies prompting strategies for text-conditioned forecasting~\citep{ashok2025beyond}. However, language pretraining does not provide the temporal inductive biases used by specialised forecasting models~\citep{tan2024useful}. LLMs also show limited performance on controlled tests of numerical time-series reasoning~\citep{merrill2024}. Instead of asking the LLM to generate forecasts, \rc{} uses it to rank TSFM-generated candidates and evaluate complete trajectories.

\paragraph{LLM-guided planning and search.}
MCTS combines a model of state transitions, a policy for selecting actions, and a value function for evaluating trajectories. AlphaGo and AlphaZero showed that these components can be integrated through PUCT to guide search over large decision spaces~\citep{silver2017mastering}.
Recent methods have also used tree search for language-model reasoning and code generation. Tree of Thoughts~\citep{yao2023tree} searches over intermediate reasoning steps generated and evaluated by an LLM. RAP~\citep{hao2023reasoning} treats reasoning as planning with the LLM acting as both the simulator and the reasoning agent. rStar-Math~\citep{wang2024rstar} combines MCTS with process-level evaluation for mathematical reasoning. In these methods, both state generation and evaluation occur primarily in language space. \rc{} instead separates these roles across modalities: a numerical time-series model generates candidate forecast transitions, while the LLM supplies context-dependent policy and value signals to guide the search.

\section{Method}
\label{sec:method}

\begin{figure*}[t]
\centering
\includegraphics[width=\linewidth]{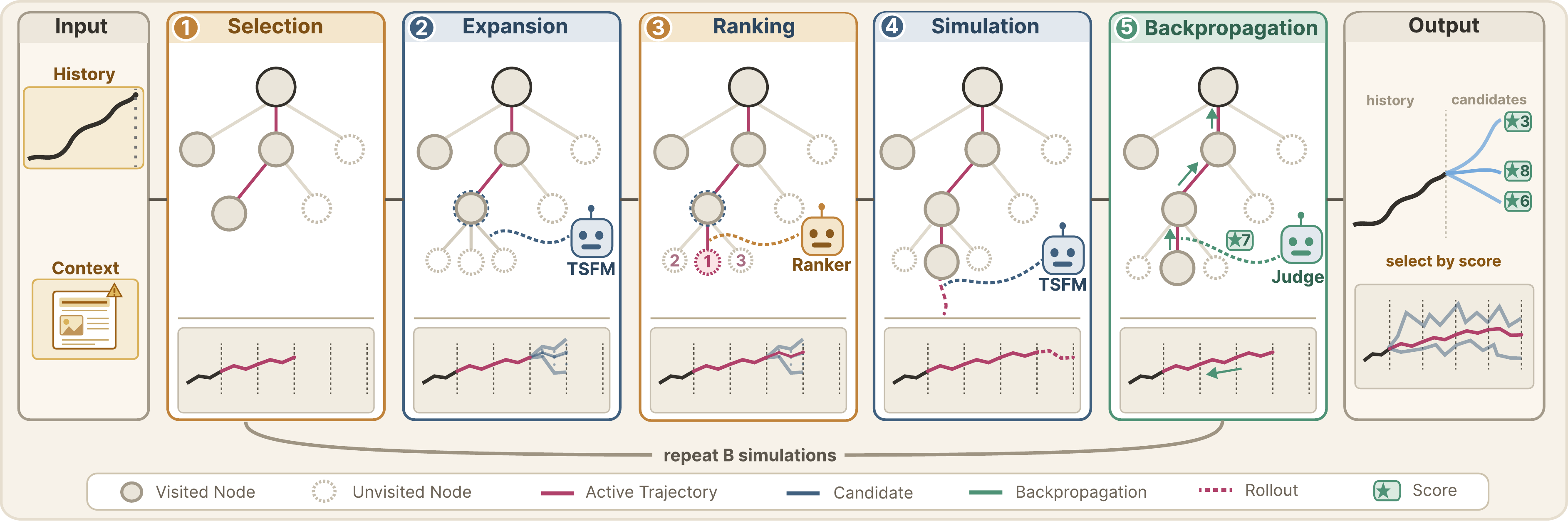}
\caption{\textbf{\rc{} overview.}
The frozen TSFM serves as the simulator, generating candidate numerical continuations and completing forecast trajectories. The LLM provides both the policy and value signals: the \emph{Ranker} assesses the relevance of the context and ranks candidate continuations, while the \emph{Judge} evaluates completed trajectories. Each of the $B$ simulations proceeds through Selection, Expansion, Ranking, Simulation, and Backpropagation. The final forecast is read from the accumulated search state, and all numerical values are generated through the frozen TSFM interface.}
\label{fig:pipeline}
\end{figure*}

\subsection{Problem formulation and overview}
\label{sec:pipeline-overview}
\label{sec:method-principles}

We formulate text-conditioned forecasting as a finite-horizon sequential decision problem. Given a numerical history $x$ and textual context $c$, the state at step $t$ is $s_t=(y_{<t},c)$, where $y_{<t}$ is the forecast prefix committed so far. An action selects one $W$-step continuation from a set of candidates generated by the frozen TSFM. The TSFM therefore acts as the \emph{simulator}, defining the numerical transitions available from the current prefix. The Ranker LLM acts as the \emph{policy}, providing context-dependent guidance over these transitions, while the Judge LLM acts as the \emph{value function}, evaluating completed forecast trajectories against the context.

When the context contains information relevant to the current forecast window, the corresponding state is treated as a \emph{decision node}, and the Ranker provides a prior over its candidate continuations. When the context provides no relevant information, the state is treated as a \emph{chance node}, and its outgoing transition is sampled uniformly from the TSFM candidates. For a horizon of length $H$ divided into $D=\lceil H/W\rceil$  windows, with $K$ candidates generated at each window, the resulting tree contains up to $K^D$ root-to-leaf trajectories. The objective is to identify high-value trajectories without exhaustively evaluating this space.

\paragraph{Pipeline overview.}
\rc{} explores the trajectory space using $B$ MCTS simulations (Figure~\ref{fig:pipeline}, Algorithm~\ref{alg:recast}). Each simulation follows five stages. \textbf{Selection} traverses the existing tree to an expandable leaf. \textbf{Expansion} uses the TSFM to generate $K$ continuations from the current prefix. \textbf{Ranking} determines whether the context is relevant to the current window and, when relevant, converts the Ranker's ordering into policy priors. \textbf{Simulation} extends the selected continuation to the end of the horizon using the TSFM. \textbf{Backpropagation} evaluates the completed trajectory with the Judge and propagates its score through the visited path. Later simulations can therefore reuse earlier evaluations to direct exploration towards promising continuations.

All numerical inputs to the LLM are serialized as text. Values are rounded and paired with timestamps, while long segments are summarized by their endpoints, range, and mean to limit prompt length. The LLM neither receives nor emits raw numerical arrays.

\subsection{MCTS search procedure}
\label{sec:method-search}

\paragraph{Selection.}
\label{sec:method-select}

Each simulation begins at the root and traverses previously expanded nodes until it reaches an expandable leaf. The transition rule depends on whether the current node was classified as a decision node or a chance node when it was expanded.

At a \emph{decision node}, the next child is selected using the PUCT rule
\begin{equation}
a^\star = \arg\max_a \frac{W_{\mathrm{sum}}(s,a)}{N(s,a)} + c_{\mathrm{puct}}\, \pi(a\mid s)\, \frac{\sqrt{\max\{1,\sum_b N(s,b)\}}}{1+N(s,a)},
\label{eq:puct}
\end{equation}
where $s$ is the current node, $W_{\mathrm{sum}}(s,a)/N(s,a)$ is the mean value backed up through action $a$ at $s$, defined as $0$ when $N(s,a)=0$, and $\pi(a\mid s)$ is the policy prior provided by the Ranker when the node is expanded. For an action that has not yet been visited, the mean-value term therefore vanishes and selection is driven by the prior-weighted exploration term, whose numerator is floored at one so that the rule remains well-defined at a newly expanded node. Early in the search, this prior directs exploration towards candidates supported by the context. As more simulations are completed, backed-up Judge values can reinforce or override the initial ranking.

At a \emph{chance node}, the next child is sampled uniformly from its $K$ TSFM-generated candidates rather than selected using PUCT. Neither the Ranker prior nor the backed-up Judge values influence this local transition, preventing context that is irrelevant to the current window from steering the backbone candidates. For sample-native backbones, uniform selection among independently drawn candidates preserves the TSFM transition distribution (App.~\ref{app:backbones}). Chance nodes nevertheless remain part of the search tree and receive backed-up values. This allows earlier decision nodes to account for stochastic downstream outcomes, while later windows can still become decision nodes when the context becomes relevant. If the Ranker abstains at every window, the search dynamics reduce to ordinary TSFM sampling.

\paragraph{Expansion.}
\label{sec:method-expand}

After Selection reaches an expandable leaf with committed prefix $y_{<t}$, the frozen TSFM generates $K$ candidate continuations for the next $W$-step window:
\begin{equation}
\{(\hat y^{(k)},\ell^{(k)})\}_{k=1}^{K} = \mathrm{TSFM}(x,y_{<t};K),
\label{eq:expand}
\end{equation}
where $\hat y^{(k)}$ denotes the $k$-th candidate continuation and $\ell^{(k)}$ is its scalar backbone score. This score is used only as a fallback prior in the without-Ranker ablation (\S\ref{sec:ablations}). The generated candidates become the outgoing transitions of the newly expanded node. Each candidate commits one $W$-step block, so appending $\hat y^{(k)}$ to $y_{<t}$ produces the state associated with the corresponding child.

\paragraph{Ranking.}
\label{sec:method-rank}
\label{sec:method-gate}

The Ranker receives the textual context $c$ and serialized representations of the $K$ candidate continuations. It first returns a relevance flag
$\texttt{relevance}\in\{\texttt{Yes},\texttt{No}\}$ indicating whether the context contains information relevant to the current forecast window. If $\texttt{relevance}=\texttt{Yes}$, the Ranker also returns an ordering $\mathrm{rank}_s(\cdot)$ over the candidates of the newly expanded node $s$. The ordering is converted into policy priors for the newly expanded node $s$ as
\begin{equation}
\pi(a\mid s)=\frac{e^{-\eta\,\mathrm{rank}_s(a)}}{\sum_{a'}e^{-\eta\,\mathrm{rank}_s(a')}},\qquad\eta=1.0.
\label{eq:rank}
\end{equation}
Higher-ranked candidates therefore receive larger prior probabilities, and the expanded node is stored as a decision node. If $\texttt{relevance}=\texttt{No}$, including cases in which the Ranker response cannot be parsed, the Ranker abstains and the node is stored as a chance node. Its outgoing transitions are subsequently sampled uniformly. The relevance prediction therefore determines where contextual guidance enters the tree rather than whether the node participates in search.

\paragraph{Simulation.}
\label{sec:method-rollout}

The selected continuation is appended to the committed prefix and extended to the end of the forecasting horizon using the same frozen TSFM interface, producing a complete trajectory $\tau$. Sample-native models draw trajectories directly from their predictive distributions, whereas models that provide only quantile forecasts construct trajectories from their quantile outputs (App.~\ref{app:backbones}). The TSFM is therefore the sole numerical generator during both Expansion and Simulation. The LLM neither generates nor modifies forecast values; it only guides the search over trajectories produced through the TSFM interface. Consequently, the method remains limited by the range of futures that the backbone can generate.

\paragraph{Backpropagation.}
\label{sec:method-judge}

The Judge LLM acts as the value function by evaluating how consistent the completed trajectory $\tau$ is with the textual context $c$. It emits a single digit $\texttt{SCORE}\in\{0,\dots,9\}$. Rather than using only the greedily decoded digit, we compute the expected score under the model's distribution over the score token:
\begin{equation}
s(\tau) = \mathbb{E}[\texttt{SCORE}] = \sum_{k=0}^{9} k\,P(\texttt{SCORE}{=}k\mid c,\tau) \in[0,9].
\label{eq:judge}
\end{equation}
Here $P(\texttt{SCORE}{=}k\mid c,\tau)$ is the probability that the Judge assigns to the digit token $k$ at the output position where the score is emitted. The Judge first produces a short analysis under greedy decoding; at the score position, the next-token probabilities of the ten digit tokens $\{0,\dots,9\}$ are read and renormalized to sum to one. The distribution is therefore conditioned on the context, the serialized trajectory, and the Judge's own analysis. Taking the expectation rather than the greedily decoded digit yields a continuous value that distinguishes trajectories receiving the same discrete score.

The resulting value $s(\tau)$ is used as the MCTS reward and propagated along the committed path. For every edge $(s,a)\in\mathcal{E}(\tau)$ visited by the simulation,
\begin{equation}
N(s,a)\mathrel{+}=1, \qquad W_{\mathrm{sum}}(s,a)\mathrel{+}=s(\tau).
\end{equation}
These updates allow later simulations to use previous trajectory evaluations when selecting which branches to explore. The Judge uses greedy decoding with temperature $0$, so identical trajectory--context pairs can be cached without changing the result (\S\ref{sec:cost}).

\subsection{Output selection}
\label{sec:method-output}

After $B$ simulations, the search state contains two forms of information: the tree, whose edges $(s,a)$ store $(N(s,a),W_{\mathrm{sum}}(s,a))$, and a pool of $B$ completed rollouts with their Judge scores. Both output modes are read from this accumulated state. Each rollout $\tau_i$ is assigned a path value equal to the mean backed-up value along its committed tree path:
\begin{equation}
\mathrm{pathQ}(\tau_i)=\frac{1}{|\mathcal{E}(\tau_i)|}\sum_{(s,a)\in\mathcal{E}(\tau_i)}\frac{W_{\mathrm{sum}}(s,a)}{N(s,a)},
\label{eq:pathq}
\end{equation}
where $\mathcal{E}(\tau_i)$ is the set of tree edges on the path associated with $\tau_i$. Because the Judge score and the mean backed-up edge values both lie on the same $[0,9]$ scale, they are combined as
\begin{equation}
\mathrm{value}(\tau_i)=\frac{s(\tau_i)+\alpha\,\mathrm{pathQ}(\tau_i)}{1+\alpha},\qquad\alpha=0.5.
\label{eq:pickbest}
\end{equation}
The mixing weight $\alpha$ balances the individual Judge verdict with the value accumulated across related simulations.

For distributional evaluation on CiK, we sample 25 trajectories from the tree using visit-proportional sampling, with actions drawn according to $N(s,a)^{1/\tau}$. This matches the baseline output format. For point forecasting on Time-MMD, we use a trust-region rule. We retain the five rollouts with the highest $\mathrm{value}$ and return the shortlisted trajectory nearest the geometric median of all $B$ rollouts:
\begin{equation}
\tau^\star=\arg\min_{i\in\mathrm{top}\text{-}5}\left\|\tau_i-\mathrm{gmedian}\bigl(\{\tau_1,\dots,\tau_B\}\bigr)\right\|_2.
\end{equation}
The shortlist incorporates contextual preference, while the geometric median limits the influence of a noisy Judge score by favouring trajectories near the backbone-generated consensus. The shortlist size is fixed at five (App.~\ref{app:hyperparams}). TSFM-BoN applies the same output rules, so the comparison isolates candidate construction. All returned forecast values are generated through the frozen TSFM interface.

\section{Experimental Setup}
\label{sec:setup}

\subsection{Benchmarks}
\textbf{CiK}~\citep{williams2024cik} contains 71 tasks across seven domains, with five deterministic instances per task, yielding 355 instances in total. The benchmark covers history lengths of 3--287 timesteps and horizons of 6--99, with 14 tasks imposing explicit numerical constraints. We use the official weighted region-of-interest CRPS (RCRPS; App.~\ref{app:metrics}), which normalizes scores per task and penalizes constraint violations. Future- and covariate-source contexts provide covariates or constraints rather than future target values and therefore do not introduce label leakage.

\textbf{Time-MMD}~\citep{liu2024timemmd} contains 9 domains (Agriculture, Climate, Economy, Security, SocialGood, Traffic, monthly; Energy, Health, weekly; Environment, daily), four horizons per domain (monthly $6/8/10/12$, weekly $12/24/36/48$, daily $48/96/192/336$), $50$ windows per (domain, horizon, seed), with $24$-step conditioning (following TaTS protocol~\citep{li2026language}). We score MSE \emph{and} MAE in \texttt{StandardScaler} space and, matching the headline table (Table~\ref{tab:tats-main}), aggregate as the \emph{geometric mean over the nine domains} of the per-domain error (\S\ref{sec:tats}). Full dataset statistics are in App.~\ref{app:datasets}.

\subsection{Baselines}

We organize the baselines into three groups, corresponding to the row groups in Table~\ref{tab:cik-main}. The groups are ordered by the degree of control given to the LLM over the forecast values. \rc{} forms a fourth group, \emph{LLMs as forecasting planners}.

\begin{itemize}
\item \textbf{TSFM only.}
\emph{Zero-shot TSFM}: the frozen backbone is used without textual context. On CiK, it produces 25 stochastic samples; on Time-MMD, it produces one sample for each forecast window.

\item \textbf{LLMs as forecasters.}
These methods ask the LLM to generate the forecast values directly, without using a TSFM backbone, and are therefore backbone-independent. \emph{DirectPrompt} uses zero-shot prompting to generate forecast values from the numerical history and textual context. \emph{LLM-CoT} \citep{wei2022chain} adds a chain-of-thought reasoning step before producing the numerical forecast. \emph{NEXUS}~\citep{das2026nexus} uses multi-agent synthesis and is evaluated with the same open-source LLMs.

\item \textbf{LLMs as verifiers.}
These methods use the LLM only to score candidates generated by the TSFM. \emph{TSFM-BoN} performs flat best-of-$N$ reranking: it samples $N$ candidates from the backbone, scores each candidate using the same continuous Judge as \rc{}, and then applies the same output rule without tree search. On CiK, it returns the Judge's top-25 trajectories as an ensemble. On Time-MMD, it applies the same trust-region rule as \rc{}. Thus, \rc{} and TSFM-BoN differ only in how their candidate pools are constructed. 
\end{itemize}

\subsection{Implementation}
\label{sec:impl}

\textbf{Backbones.}
\rc{} uses each TSFM as a frozen backbone without updating its parameters. The main experiments (Table~\ref{tab:cik-main}) include two architecturally distinct models: Chronos-T5-Large~\citep{ansari2024chronos}, an autoregressive token model that produces joint trajectory samples and serves as our default backbone; and TimesFM-1.0-200M~\citep{das2024timesfm}, a quantile-based model whose samples are reconstructed from its per-step quantile predictions (see details in App.~\ref{app:backbones}). 

\textbf{Ranker and Judge LLMs.}
The Ranker and Judge are served using vLLM~\citep{kwon2023vllm} on one V100 GPU, while the frozen TSFM runs on a second GPU (Table~\ref{tab:cost}). To evaluate sensitivity to the choice of LLM, the main experiments use four instruction-tuned models: Qwen2.5-3B, Qwen2.5-7B, Llama-3.1-8B, and Gemma-3-4B. The full Ranker and Judge prompts are listed in App.~\ref{app:prompts}.

\textbf{Hyperparameters.}
The default configuration samples $K=8$ candidates at each expanded node, runs $B=64$ MCTS simulations, and limits the search depth to $D=8$. All reported \rc{} results use this setting. To ensure a fair comparison, TSFM-BoN evaluates $64$ complete trajectories, matching the simulation budget and the number of Judge calls used by \rc{}, and applies the same final output rule. Thus, the comparison isolates how the evaluation budget is allocated rather than differences in Judge usage or output selection. For CiK, \rc{} returns 25 trajectories for distributional evaluation, whereas for Time-MMD it returns a single trajectory using the trust-region rule described in \S\ref{sec:method-output}. Complete hyperparameter settings are provided in App.~\ref{app:hyperparams}.

\section{Main Results}
\label{sec:results}
\label{sec:cik}
\label{sec:tats}

\begin{table*}[t]
\centering
\caption{Main results on CiK (weighted RCRPS $\downarrow$, left) and Time-MMD (geometric-mean MSE $\downarrow$ over the nine domains, right), averaged over 3 seeds, per Ranker/Judge LLM $\times$ TSFM (\emph{Chr}onos-T5-Large, \emph{Tim}esFM-1.0-200M). Best score per column in \textbf{bold}; per-domain Time-MMD tables in App.~\ref{app:timemmd-perdataset}; per-seed standard deviations in App.~\ref{app:std}.}
\label{tab:cik-main}\label{tab:tats-main}
\scriptsize
\setlength{\tabcolsep}{2.5pt}
\resizebox{\textwidth}{!}{%
\begin{tabular}{@{}l cccccccc cccccccc@{}}
\toprule
& \multicolumn{8}{c}{CiK weighted RCRPS $\downarrow$}
& \multicolumn{8}{c}{Time-MMD geo-mean MSE $\downarrow$} \\
\cmidrule(lr){2-9}\cmidrule(lr){10-17}
& \multicolumn{2}{c}{Qwen2.5-3B} & \multicolumn{2}{c}{Qwen2.5-7B}
& \multicolumn{2}{c}{Llama-3.1-8B} & \multicolumn{2}{c}{gemma-3-4B}
& \multicolumn{2}{c}{Qwen2.5-3B} & \multicolumn{2}{c}{Qwen2.5-7B}
& \multicolumn{2}{c}{Llama-3.1-8B} & \multicolumn{2}{c}{gemma-3-4B} \\
\cmidrule(lr){2-3}\cmidrule(lr){4-5}\cmidrule(lr){6-7}\cmidrule(lr){8-9}
\cmidrule(lr){10-11}\cmidrule(lr){12-13}\cmidrule(lr){14-15}\cmidrule(lr){16-17}
Method
  & Chr. & Tim. & Chr. & Tim. & Chr. & Tim. & Chr. & Tim.
  & Chr. & Tim. & Chr. & Tim. & Chr. & Tim. & Chr. & Tim. \\
\midrule
\multicolumn{17}{@{}l}{\textit{TSFM only}} \\
Zero-shot TSFM
  & 0.274 & 0.254 & 0.274 & 0.254 & 0.274 & 0.254 & 0.274 & 0.254
  & 1.883 & 0.863 & 1.883 & 0.863 & 1.883 & 0.863 & 1.883 & 0.863 \\
\midrule
\multicolumn{17}{@{}l}{\textit{LLMs as forecaster}} \\
DirectPrompt
  & \multicolumn{2}{c}{0.474} & \multicolumn{2}{c}{0.267} & \multicolumn{2}{c}{0.478} & \multicolumn{2}{c}{0.784}
  & \multicolumn{2}{c}{4.219} & \multicolumn{2}{c}{1.513} & \multicolumn{2}{c}{4.284} & \multicolumn{2}{c}{1.458} \\
LLM-CoT
  & \multicolumn{2}{c}{0.322} & \multicolumn{2}{c}{\textbf{0.220}} & \multicolumn{2}{c}{0.464} & \multicolumn{2}{c}{0.593}
  & \multicolumn{2}{c}{331.503} & \multicolumn{2}{c}{2.603} & \multicolumn{2}{c}{606.239} & \multicolumn{2}{c}{9.353} \\
NEXUS
  & \multicolumn{2}{c}{0.380} & \multicolumn{2}{c}{0.344} & \multicolumn{2}{c}{0.493} & \multicolumn{2}{c}{0.423}
  & \multicolumn{2}{c}{0.927} & \multicolumn{2}{c}{0.929} & \multicolumn{2}{c}{1.103} & \multicolumn{2}{c}{\textbf{0.821}} \\
\midrule
\multicolumn{17}{@{}l}{\textit{LLMs as verifier}} \\
TSFM-BoN
  & 0.274 & 0.262 & 0.277 & 0.256 & 0.277 & 0.264 & 0.292 & 0.263
  & 0.782 & 0.773 & 0.754 & 0.782 & 0.781 & \textbf{0.771} & 1.389 & 0.801 \\
\midrule
\multicolumn{17}{@{}l}{\textit{LLMs as forecasting planner}} \\
\textbf{\rc{} (ours)}
  & \textbf{0.262} & \textbf{0.250} & 0.260 & 0.245 & \textbf{0.258} & \textbf{0.251} & \textbf{0.264} & \textbf{0.245}
  & \textbf{0.768} & \textbf{0.769} & \textbf{0.744} & \textbf{0.772} & \textbf{0.738} & 0.772 & 1.387 & \textbf{0.787} \\
\bottomrule
\end{tabular}%
}
\end{table*}

\textbf{Comparison with the context-blind anchor.}
Table~\ref{tab:cik-main} evaluates a 25-trajectory predictive distribution on CiK using weighted RCRPS and a single selected trajectory on Time-MMD using geometric-mean MSE. Across both benchmarks, \rc{} matches or improves on the corresponding context-blind TSFM in all 16 LLM--backbone configurations, reducing error by up to approximately $4\%$ on CiK and $60\%$ on Time-MMD. The baselines do not exhibit the same consistency. Flat reranking remains worse than both context-blind backbones on CiK, while the LLM-as-forecaster methods underperform the corresponding backbone in most configurations. The key result is therefore the consistent improvement across benchmarks, LLMs, and TSFM backbones rather than the largest gain in any individual setting.

\textbf{Comparison with LLM-as-forecaster baselines.}
LLMs as forecaster methods are highly sensitive to the choice of LLM. On CiK, performance for the same prompting method varies by approximately $2.7$-$3\times$ across LLMs; on Time-MMD, it ranges from competitive with the context-blind anchors to errors several orders of magnitude larger. These methods can nevertheless achieve strong individual results. LLM-CoT with Qwen2.5-7B obtains the best CiK score, reducing error by approximately $13$-$20\%$ relative to the two anchors, which suggests that CiK's curated and explicit contexts can support direct forecasting by a capable LLM. Similarly, NEXUS outperforms the Chronos anchor for every LLM on Time-MMD, although it improves on the stronger TimesFM-1 anchor in only one configuration. These gains are not stable across models or prompting strategies: LLM-CoT with the other LLMs performs $27$-$134\%$ worse than the stronger CiK anchor, while DirectPrompt with Qwen2.5-7B is approximately $5\%$ worse. In contrast, by separating contextual reasoning from numerical generation, \rc{} maintains substantially more stable performance across the same LLM choices.

\textbf{Comparison with Best-of-$N$ baseline.} 
TSFM-BoN uses the same TSFM backbone, Judge, number of complete TSFM rollouts, and final output rule as \rc{}, but evaluates a fixed pool of complete trajectories rather than searching over partial forecasts. Both methods use $64$ Judge calls, providing a matched terminal-evaluation budget, although \rc{} additionally invokes the Ranker to guide the search. On CiK, TSFM-BoN remains worse than both zero-shot TSFMs, whereas \rc{} improves on them and reduces error by approximately $2$--$10\%$ relative to TSFM-BoN, depending on the backbone. On Time-MMD, TSFM-BoN improves on both zero-shot TSFMs in every configuration, showing that verification alone is effective. Nevertheless, \rc{} performs better in seven of the eight LLM--backbone combinations, and its best configuration for each backbone outperforms the corresponding best TSFM-BoN configuration. The only exception is TimesFM-1 with Llama-3.1-8B, for which TSFM-BoN is marginally better. This isolated reversal is consistent with TimesFM-1's less diverse quantile-reconstructed candidate pool and the weak relationship between Judge scores and forecast error observed in Fig.~\ref{fig:mechanism}b. When candidates are similar and the Judge provides limited discrimination among them, sequentially reallocating simulations offers little advantage over flat reranking and may slightly amplify ranking noise.

\begin{figure*}[t]
\centering
\includegraphics[width=\textwidth]{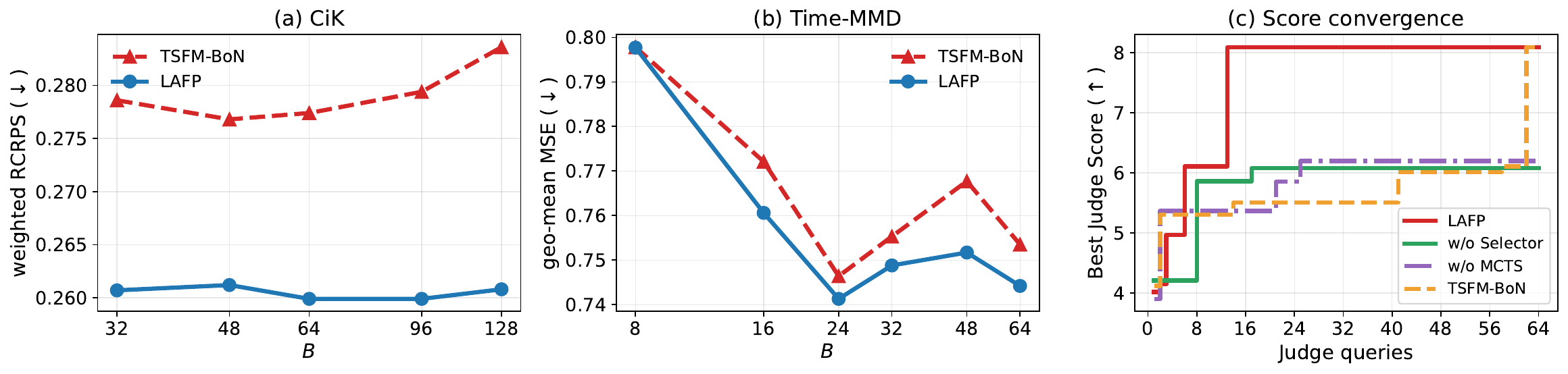}
\caption{Budget curve, Chronos-T5-Large TSFM with Qwen2.5-7B Ranker/Judge. \textbf{(a)} CiK (weighted RCRPS) and \textbf{(b)} Time-MMD (geo-mean MSE): \rc{} vs.\ TSFM-BoN at matched Judge-call budget $B$. \textbf{(c)} Best-so-far Judge score vs.\ Judge queries.}
\label{fig:budget}
\end{figure*}

\textbf{Effect of the TSFM backbone.}
\rc{} improves over the corresponding context-blind forecast with both backbones, although the magnitude of the gain depends on how each backbone generates candidate trajectories. On CiK, \rc{} improves on the context-blind anchor by approximately $4$--$6\%$ with Chronos and $1$--$4\%$ with TimesFM-1. The difference is larger on Time-MMD. Chronos generates diverse joint trajectories, giving the search greater scope to identify substantially better continuations and reducing error by up to $60\%$. The trust-region output rule also stabilizes its more variable samples (App.~\ref{app:timemmd-perdataset}). TimesFM-1 reconstructs trajectories from per-step quantiles and therefore produces a narrower, less diverse candidate pool. Its gains are smaller, at approximately $10\%$, but more consistent across LLM configurations. These results show that \rc{} benefits both backbone families, while the diversity and dependence structure of their samples determine how much sequential search can improve the forecast.

\textbf{Effect of the LLM.}
On CiK, \rc{} is relatively robust to the choice of Ranker and Judge. Changing the LLM alters the score by approximately $2\%$ with either backbone, substantially less than the variation observed for the LLM-as-forecaster baselines. Even Qwen2.5-3B captures most of the improvement, suggesting that ranking and evaluating TSFM-generated trajectories is less demanding than generating forecast values directly. The choice of LLM has a larger effect on Time-MMD, where the method must select a single trajectory. With Chronos, the strongest configurations reduce error by approximately $60\%$, whereas the weakest-performing LLM, Gemma-3-4B, achieves a reduction of about $26\%$ and is the only \rc{} configuration outperformed by NEXUS. This result indicates that using a weaker LLM can limit the benefit of planning, particularly when selecting a single forecast from a diverse candidate pool. Nevertheless, every LLM-backbone configuration matches or improves on its corresponding context-blind TSFM.

\textbf{Failure modes of direct numerical generation.}
The per-domain results in App.~\ref{app:timemmd-perdataset} reveal several extreme failures in Security, Health, and Environment, primarily for weaker LLM-as-forecaster configurations. As detailed in App.~\ref{app:timemmd-divergence}, these failures arise mainly from incorrect scale or unit assumptions rather than incorrect trends. For example, Security targets can reach hundreds of millions despite the text providing no numerical scale, while Health and Environment reports may leave LLMs to confuse percentages, such as $0$-$8\%$, with absolute counts. \rc{} avoids this failure mode because the TSFM generates all numerical values and the LLM only evaluates their contextual consistency; ambiguity in the text therefore cannot directly determine the forecast scale.

\section{Ablations and Analysis}
\label{sec:ablations}
\subsection{Ablations}
\label{sec:ablation-textmod}

\begin{table}
\centering
\caption{Component ablations with Qwen2.5-7B as Ranker/Judge. \emph{w/o Ranker} replaces the context-based prior with the TSFM score, so the search receives no contextual guidance; \emph{w/o MCTS} scores independently sampled complete trajectories at the same Judge-call budget, without selection or value backup. Best per backbone and column in \textbf{bold}.}
\label{tab:components}
\setlength{\tabcolsep}{6pt}
\begin{tabular}{@{}lcc@{}}
\toprule
Configuration & CiK RCRPS $\downarrow$ & Time-MMD geo.MSE $\downarrow$ \\
\midrule
\multicolumn{3}{l}{\textit{Chronos-T5-Large}} \\
\quad Full \rc{} & \textbf{0.260} & \textbf{0.744} \\
\quad w/o Ranker & 0.284 & 0.946 \\
\quad w/o MCTS & 0.269 & 0.764 \\
\midrule
\multicolumn{3}{l}{\textit{TimesFM-1.0-200M}} \\
\quad Full \rc{} & \textbf{0.245} & 0.772 \\
\quad w/o Ranker & 0.260 & \textbf{0.751} \\
\quad w/o MCTS & 0.251 & 0.782 \\
\bottomrule
\end{tabular}
\end{table}

On CiK, both the Ranker and MCTS contribute consistently, with the full method achieving the best result for both backbones. Removing the Ranker increases RCRPS by approximately $6$-$7\%$, while replacing MCTS with flat sampling at the same budget increases it by about $2\%$. On Time-MMD, MCTS also improves both backbones, reducing error by approximately $2\%$ for Chronos and $1\%$ for TimesFM-1. The effect of the Ranker is less stable. Removing it increases the Chronos error by about $27\%$ but improves TimesFM-1 by approximately $3\%$. This difference may arise from the interaction between the benchmark and the backbone. CiK provides curated contexts with explicit information for distinguishing among candidate futures and evaluates a predictive distribution. Time-MMD instead provides shorter reports and evaluates a single selected trajectory, making the result more sensitive to ranking noise. This is particularly relevant for TimesFM-1, whose relatively homogeneous candidate pool provides limited separation for the Ranker. We therefore do not claim that the Ranker consistently improves point-forecast accuracy on Time-MMD.

\begin{figure*}[t]
\centering
\includegraphics[width=\textwidth]{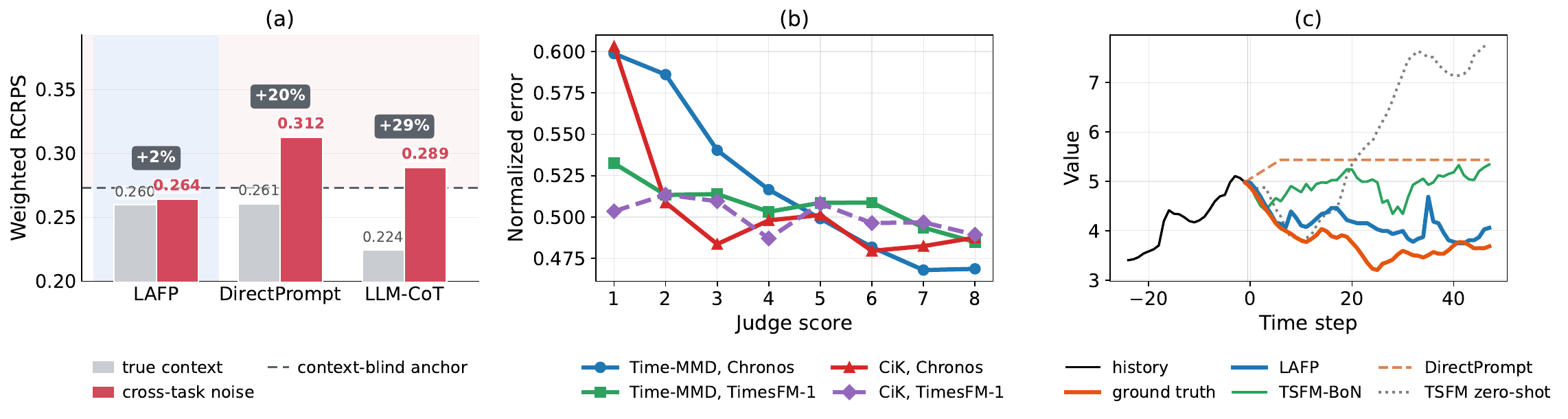}
\caption{\textbf{(a)} CiK weighted RCRPS, true vs.\ cross-instance shuffled context; dashed line: context-blind anchor. \textbf{(b)} Mean within-window error percentile vs.\ Judge score. \textbf{(c)} Forecast comparison of \rc{} and baselines.}
\label{fig:mechanism}
\end{figure*}

\subsection{Analysis}
\label{sec:budget}
\label{sec:error-analysis}


\textbf{Budget curves.}
Fig.~\ref{fig:budget}a,b traces both methods over the simulation budget $B$ under matched TSFM rollouts, Judge calls, and output rules. The two benchmarks show different regimes. On Time-MMD, both methods improve as the budget grows, and most of the improvement is realized within the first third of the default budget; \rc{} stays at or below TSFM-BoN at every evaluated budget, by up to approximately $2\%$ (Fig.~\ref{fig:budget}b). On CiK, \rc{} is essentially flat across the full budget range, whereas TSFM-BoN degrades as the pool grows, so the gap widens from approximately $6\%$ at the default budget to $8\%$ at twice the default (Fig.~\ref{fig:budget}a). Because the two methods score the same rollout pool, this divergence isolates how the pool is read out. Returning the Judge's top-scored trajectories becomes more error-prone as a larger pool contains more over-scored outliers, whereas the visit-based read-off of \rc{} concentrates on branches whose backed-up values remain high across many simulations, and is therefore stable as the pool grows. Curves for all eight Ranker/Judge LLM $\times$ TSFM pairs are shown in App.~\ref{app:budget-pairs}.

\textbf{Judge score convergence.} Figure~\ref{fig:budget}c illustrates this behaviour on a CiK instance in which the relevant information is embedded among distractor text. \rc{} reaches the highest Judge score in the candidate pool using nearly $5\times$ fewer queries than TSFM-BoN. The flat reranker allocates the same number of Judge calls without using previous scores to guide subsequent sampling. Neither ablation reaches the same score. Without the Ranker, the search rarely explores the relevant branch; without value backup, it cannot allocate additional simulations to that branch after it is found. The Judge score is an internal search objective rather than a direct measure of forecast accuracy. However, the result shows how selection and backup reuse earlier evaluations to guide later queries, which explains the average gains in Fig.~\ref{fig:budget}a,b.

\textbf{Text robustness.}
In Fig.~\ref{fig:mechanism}a, each CiK context is augmented with real sentences drawn from other tasks, while the time series and original context remain unchanged. The methods therefore receive relevant information mixed with fluent but misleading cross-task content. DirectPrompt degrades by approximately $20\%$ and LLM-CoT by about $29\%$, leaving both methods $6$-$14\%$ worse than the context-blind anchor. In contrast, \rc{} changes by only about $2\%$ and remains better than the anchor. This robustness follows from the restricted role of the LLM. In the forecaster baselines, corrupted text directly affects the generated numerical values. In \rc{}, the LLM only ranks and evaluates TSFM-generated trajectories, so misleading context can steer the search towards weaker candidates but cannot directly alter the forecast values. Its effect is therefore bounded by the candidate support of the backbone.

\textbf{Judge analysis.}
Figure~\ref{fig:mechanism}b groups rollouts by Judge score and reports their forecasting errors as rank percentiles within the corresponding candidate set. For Chronos, error percentiles decrease as Judge scores increase on both benchmarks, indicating that the Judge reliably assigns lower scores to poorer trajectories. For TimesFM-1, the curves remain near $0.5$, suggesting a weaker association between Judge score and forecasting error. This difference is consistent with the backbones' sampling behaviour. Chronos generates diverse joint trajectories that include more extreme failures, whereas TimesFM-1 reconstructs trajectories from per-step quantiles and produces a narrower candidate pool. Most of the separation occurs at low Judge scores, while trajectories receiving high scores remain difficult to distinguish. The Judge is therefore more effective at filtering poor trajectories than at precisely ranking the strongest ones. This behaviour aligns with \rc{}, which uses backed-up Judge values to reduce further exploration of weak branches and derives the final forecast from the accumulated search state rather than selecting solely by the highest Judge score. It also helps explain why replacing MCTS with flat sampling has a larger effect for Chronos than for TimesFM-1 in Table~\ref{tab:components}.

\textbf{Prediction visualization.}
Figure~\ref{fig:mechanism}c compares \rc{} with the baselines on a Time-MMD Energy window with horizon $H=48$. The context-blind forecast deviates substantially from this trajectory, while DirectPrompt fails to recover the lower post-reversal level. TSFM-BoN captures part of the reversal but remains systematically above the ground truth across most of the horizon. In contrast, \rc{} most closely matches both the magnitude of the decline and the subsequent level, yielding a squared error several times lower than the best baseline on this instance. This example shows that guided sequential search can identify TSFM-generated trajectories that better align with the realised future than direct forecasting or flat reranking. Aggregate results are reported in Table~\ref{tab:tats-main} and Fig.~\ref{fig:budget}.

\label{sec:cost}
\begin{table}[t]
\centering
\caption{Per-instance LLM calls, TSFM full-trajectory generations, and median wall-clock (Qwen2.5-7B Ranker/Judge, $K{=}8$, $B{=}64$; one two-GPU V100 node, sequential). Italic latencies are amortized over a batched vLLM run.}
\label{tab:cost}
\small
\setlength{\tabcolsep}{4pt}
\begin{tabular}{@{}l cc c@{}}
\toprule
Method & LLM calls & TSFM traj.\ calls & s / inst.\ \\
\midrule
\multicolumn{4}{@{}l}{\textit{CiK (355 instances)}} \\
\rc{} (Chronos)    & 104.5 & 64 & 121 \\
\rc{} (TimesFM-1)  & 104.9 & 64 & 89 \\
\midrule
\multicolumn{4}{@{}l}{\textit{Time-MMD (1{,}793 windows)}} \\
\rc{} (Chronos)    & 110.9 & 64 & 98 \\
\rc{} (TimesFM-1)  & 110.9 & 64 & 81 \\
TSFM-BoN        & 64    & 64 & 21.8 \\
NEXUS              & 76    & 0  & 223 \\
DirectPrompt       & 1     & 0  & \emph{0.5} \\
LLM-CoT            & 1     & 0  & \emph{1.1} \\
\bottomrule
\end{tabular}
\end{table}

\textbf{Cost and latency.}
For each instance, \rc{} makes one Ranker call per expanded node, approximately two-thirds as many calls as the search budget $B$, exactly $B$ Judge calls, and one full-horizon TSFM rollout per simulation (Table~\ref{tab:cost}). We report call counts because they are independent of hardware, batching, and serving configuration; wall-clock latency additionally depends on the number of sequential inference rounds and the generation length of each call. TSFM-BoN evaluates its entire candidate pool in a single batched round, whereas \rc{} distributes its Judge calls across sequential search iterations. TSFM-BoN is therefore several times faster than \rc{}, by more than its lower call count alone would suggest, but this efficiency comes with the accuracy gap reported in Table~\ref{tab:cik-main}. In our reimplementation, NEXUS makes approximately $30\%$ fewer LLM calls than \rc{}, one context-structuring call followed by sampled generations from the macro, micro, and synthesizer agents, yet is more than twice as slow when using the same LLM. Its calls generate long, free-form reasoning and numerical forecasts, whereas the Ranker and Judge return short verdicts over TSFM-generated candidates. Because the Judge uses greedy decoding, repeated trajectory--context pairs could in principle be cached (\S\ref{sec:method-judge}); however, judged trajectories are almost always distinct in our experiments, so all reported counts are uncached. When lower cost or latency is required, halving the search budget increases error by approximately $3\%$ (Fig.~\ref{fig:budget}).
\section{Conclusion}
\label{sec:conclusion}
We formulate text-conditioned time-series forecasting as a planning problem in which each model is assigned a role aligned with its strengths. Rather than generating or revising forecast values, the LLM serves as a Ranker and Judge over trajectories produced by a frozen TSFM. \rc{} implements this formulation through Monte Carlo tree search. The Ranker determines when the context is relevant and guides the exploration of candidate continuations, while the Judge evaluates completed trajectories and provides values for subsequent search. Across two benchmarks, two TSFM backbones, and four LLMs, \rc{} matches or improves on the corresponding context-blind TSFM in every configuration and generally outperforms flat Best-of-$N$ reranking under matched Judge-call and rollout budgets. These results show that sequentially reusing earlier evaluations is more effective than independently scoring a fixed pool of complete trajectories. Direct LLM forecasting can achieve strong results in selected settings, but is substantially more sensitive to the choice of model and to corrupted context. \rc{} remains limited by the futures that its backbone can generate and by the quality of the LLM guidance. More broadly, the framework demonstrates how heterogeneous foundation models can be composed without training by restricting each model to decisions supported by its native capabilities.

\bibliographystyle{plainnat}
\bibliography{references}

\clearpage
\appendix
\section*{Appendix}

\section{Dataset Statistics}
\label{app:datasets}

\subsection{CiK}
CiK~\cite{williams2024cik} comprises 71 tasks across seven application domains; 95\% of the tasks use real-world data, and the three simulated SVAR tasks are grouped under Mechanics alongside the causal-chamber tasks. We follow the benchmark's deterministic 5-instance selection (355 instances) and its per-task weighting (\texttt{TASK\_NAME\_TO\_WEIGHT}), which rebalances the aggregate away from raw task counts (e.g., Public Safety contributes 26/71 tasks but only 3/21 of the total weight). Table~\ref{tab:cik-stats} is computed directly from the 355 evaluation instances. History lengths span 3--287 steps and horizons 6--99: the Public Safety and Economics tasks are deliberately history-starved (3--6 observations), making the textual context the dominant information source, whereas Transportation, Retail, and Energy provide 96--168-step histories with strong daily/weekly cycles. Sampling frequencies range from seconds (Mechanics) through 10-minute/hourly (Climatology, Transportation, Energy) and daily (Retail) to monthly (Public Safety, Economics).

\begin{table}[h]
\centering
\caption{CiK domain summary, computed from the benchmark's 355 evaluation instances. Hist.\ / $H$ = history and horizon length ranges (timesteps). Context sources~\cite{williams2024cik}: i = intemporal, h = historical, f = future, cov = covariate, ca = causal. Constr.\ = tasks with an explicit numerical constraint (RCRPS adds a violation penalty). Wt.\ = the domain's share of the benchmark's task weighting.}
\label{tab:cik-stats}
\small
\setlength{\tabcolsep}{3pt}
\begin{tabular}{@{}lrlllcc@{}}
\toprule
Domain & \#Tasks & Hist. & $H$ & Ctx.\ sources & Constr. & Wt. \\
\midrule
Public Safety & 26 & 3--6 & 6--7 & i,h,cov,ca & 0 & 3/21 \\
Climatology & 12 & 45--79 & 24--99 & i,h,cov & 9 & 4/21 \\
Transportation & 11 & 168 & 24--72 & i,f,cov & 2 & 4/21 \\
Energy & 7 & 96--144 & 24 & f,cov & 0 & 3/21 \\
Retail & 6 & 112--168 & 56 & i,f,cov & 0 & 3/21 \\
Mechanics & 6 & 54--287 & 15--93 & i,cov,ca & 3 & 3/21 \\
Economics & 3 & 6 & 6 & cov & 0 & 1/21 \\
\midrule
Total & 71 & 3--287 & 6--99 & -- & 14 & 21/21 \\
\bottomrule
\end{tabular}
\end{table}

\subsection{Time-MMD}
Our Time-MMD forecasting protocol uses the Time-MMD dataset~\cite{liu2024timemmd}, which contains nine domains and four forecast horizons per domain. Each domain consists of a single univariate target series paired with timestep-level textual reports. Following TaTS~\cite{li2026language}, we adopt a chronological 70/10/20 train/validation/test split, fit the evaluation \texttt{StandardScaler} using only the training split, and construct stride-1 test windows from the final $0.2n{+}24$ timesteps, using a conditioning length of 24. We evaluate 50 windows for each domain-horizon pair and seed. When fewer than 50 valid windows exist, we use the full available set; for example, Security provides 48 windows at $H{=}12$. We also exclude five SocialGood windows that have missing ground-truth values across all methods, resulting in 1{,}793 evaluated windows per seed. Table~\ref{tab:tats-stats} reports the statistics for each domain. Text coverage denotes the proportion of timesteps with a non-empty report after applying the benchmark's fact--prediction split; only the factual component is provided as conditioning context.
\begin{table}[h]
\centering
\caption{Time-MMD domain summary. Length = series length and Test = test-split length (timesteps); Text = fraction of timesteps with a non-empty textual report.}
\label{tab:tats-stats}
\small
\setlength{\tabcolsep}{3pt}
\begin{tabular}{@{}llrlrlr@{}}
\toprule
Domain & Freq.\ & Length & Span & Test & Horizons & Text \\
\midrule
Agriculture & monthly & 496 & 1983--2024 & 99 & 6/8/10/12 & 80.8\% \\
Climate & monthly & 496 & 1983--2024 & 99 & 6/8/10/12 & 100\% \\
Economy & monthly & 423 & 1989--2024 & 84 & 6/8/10/12 & 100\% \\
Security & monthly & 297 & 1999--2024 & 59 & 6/8/10/12 & 100\% \\
SocialGood & monthly & 900 & 1950--2024 & 180 & 6/8/10/12 & 59.2\% \\
Traffic & monthly & 531 & 1980--2024 & 106 & 6/8/10/12 & 100\% \\
Energy & weekly & 1479 & 1996--2024 & 295 & 12/24/36/48 & 99.3\% \\
Health & weekly & 1389 & 1997--2024 & 277 & 12/24/36/48 & 99.9\% \\
Environment & daily & 15248 & 1982--2023 & 3049 & 48/96/192/336 & 96.9\% \\
\bottomrule
\end{tabular}
\end{table}

\section{TSFM Backbones}
\label{app:backbones}

\rc{} wraps two architecturally distinct frozen backbones through a common interface. The framework requires only: (i) sampling $K$ candidate blocks of width $W$, each with a scalar log-likelihood score, and (ii) a \texttt{predict}-style rollout to the full horizon $H$. How these candidates are generated determines whether uniform selection corresponds to an unbiased sample from the backbone distribution $p_\theta$.

\paragraph{Chronos-T5-Large (token-based, sample-native).}
Autoregressive token decoding produces $K$ i.i.d.\ samples at temperature $T$, together with their joint log-probabilities. Chronos is the only backbone for which we claim the exact property that abstention recovers sampling from $p_\theta$. Accordingly, the context-blind reference on the uninformative-context slice is sampled using the same EXPAND procedure with $T{=}1.5$, rather than the native \texttt{predict()} configuration with $T{=}1.0$.

\paragraph{TimesFM-1.0-200M (patch decoder, decile interface).}
TimesFM is a decoder-only patch model that outputs nine marginal deciles,
$\{q_{0.1},\ldots,q_{0.9}\}$, at each forecast step rather than joint sample paths. Following DEFT, we synthesize candidates through per-step inverse-CDF interpolation. Specifically, for each sample and forecast step, we draw
$u_h \sim \mathcal{U}(0.1,0.9)$ and compute
\[
\widehat{y}_h
=
\operatorname{interp}
\!\left(
u_h;\{q_{0.1,h},\ldots,q_{0.9,h}\}
\right).
\]
Because the draws are independent across forecast steps, the resulting trajectories do not preserve temporal correlation.

For use as a PUCT prior, we assign each candidate a Gaussian proxy log-likelihood derived from the predicted quantiles:
\[
\mu_h = q_{0.5,h},
\qquad
\sigma_h =
\frac{q_{0.9,h}-q_{0.1,h}}{2z_{0.9}},
\qquad
\ell(\widehat{\mathbf{y}})
=
\sum_h
\log
\mathcal{N}
\!\left(
\widehat{y}_h;\mu_h,\sigma_h^2
\right),
\]
where $z_{0.9}$ is the $0.9$ quantile of the standard normal distribution. This score should be interpreted as a typicality proxy rather than an exact joint likelihood. Uniform candidate selection therefore produces a product-of-marginals sample, not an unbiased draw from a joint $p_\theta$. TimesFM-1 is consequently included as evidence of accuracy and backbone generality, but we do not claim the exact abstention-recovers-$p_\theta$ property for this backbone.

\begin{table}[h]
\centering
\caption{Backbone checkpoints used in the experiments.}
\label{tab:backbones}
\small
\setlength{\tabcolsep}{3pt}
\begin{tabular}{@{}llc@{}}
\toprule
Backbone & Checkpoint & Draw  \\
\midrule
Chronos-T5-Large & \texttt{amazon/chronos-t5-large} & token  \\
TimesFM-1.0-200M & \texttt{google/timesfm-1.0-200m-pytorch} & decile  \\
\bottomrule
\end{tabular}
\end{table}

\section{Baseline Details}
\label{app:baselines}
\begin{itemize}
\item \textbf{Zero-shot TSFM}: the frozen backbone run context-blind; CiK uses $25$ stochastic samples, Time-MMD one sample per window. Each backbone is compared against its own anchor.
\item \textbf{DirectPrompt}: zero-shot prompting that generates forecast values directly from the context and numeric history, with constrained numeric decoding and no TSFM backbone; run with all four open LLMs.
\item \textbf{LLM-CoT}: the same prompting augmented with a chain-of-thought reasoning step before the numeric output; same LLMs and parsing rules.
\item \textbf{NEXUS}~\cite{das2026nexus}: multi-agent synthesis; our matched-LLM reimplementation, run with the same four open LLMs on both benchmarks.
\item \textbf{TSFM-BoN}: flat best-of-$N$ reranking of TSFM samples by the same continuous Judge, with the \emph{matched} output rule: the Judge's top-$25$ ensemble on CiK and the same trust-region pick as \rc{} on Time-MMD. This is the matched-Judge, matched-backbone, matched-output control that isolates the search structure from flat reranking.
\end{itemize}
The full DirectPrompt and LLM-CoT prompts are listed in App.~\ref{app:prompts}.

\section{Notation and Hyperparameters}
\label{app:hyperparams}
Table~\ref{tab:notation} collects the symbols used in \S\ref{sec:method} and throughout the paper.

\begin{table*}[h]
\centering
\caption{Notation.}
\label{tab:notation}
\small
\begin{tabular}{@{}ll@{}}
\toprule
Symbol & Meaning \\
\midrule
$x$ & numeric history \\
$c$ & textual context \\
$H$ & forecast horizon (prediction length) \\
$y$ & future values to forecast \\
$p_\theta$ & frozen TSFM predictive distribution (backbone) \\
$\phi$ & frozen LLM (Ranker $+$ Judge) \\
$K$ & branching factor: candidate continuations per node \\
$B$ & Monte-Carlo simulation budget \\
$D,\,W$ & number of forecast blocks ($D=\lceil H/W\rceil$) and block width \\
$s$ & search node: committed prefix and context, $s=(y_{<t},c)$ \\
$a,\,N(s,a),\,W_{\text{sum}}(s,a)$ & action; visit count and backed-up value sum of edge $(s,a)$ \\
$Q(s,a)$ & $W_{\text{sum}}(s,a)/N(s,a)$; $\mathrm{pathQ}$ averages $Q$ over a path \\
$c_{\text{puct}},\,\eta$ & PUCT constant; rank$\to$prior temperature (\texttt{prior\_temp}) \\
$T$ & backbone (Chronos) sampling temperature ($=1.5$) \\
$s(\tau)$ & continuous Judge value in $[0,9]$; $\tau$ is a trajectory (Eq.~\eqref{eq:judge}) \\
$T_{\text{vis}}$ & visit-sampling temperature (CiK read-off) \\
$\beta$ & time-attribution credit weight (\texttt{attr\_beta}) \\
$\alpha$ & pick-best blend weight (Eq.~\eqref{eq:pickbest}) \\
$k$ & shortlist width of the trust-region pick rule (\S\ref{sec:method-output}) \\
\bottomrule
\end{tabular}
\end{table*}

Table~\ref{tab:hparams} lists the reference \rc{} configuration; these are the actual settings from the method source, not placeholders.

\begin{table}[h]
\centering
\caption{\rc{} hyperparameters (reference configuration).}
\label{tab:hparams}
\small
\begin{tabular}{@{}ll|ll@{}}
\toprule
Param & Value & Param & Value \\
\midrule
$K$ (branching) & 8 & $\beta$ (attr) & 1.0 \\
$B$ (simulations) & 64 & Judge & continuous 0--9 \\
\texttt{mcts\_P} (parallel exp.) & 8 &  $\alpha$  & 0.5 \\
$E_{\text{target}}$ (max depth) & 8 & $n_{\text{samples}}$ (CiK) & 25 \\
$W_{\min}$ (block width) & 1 & pick (Time-MMD) & trust-region \\
$c_{\text{puct}}$ & 1.25 & Judge temp & 0.0 \\
$T_{\text{vis}}$ (visit temp) & 1.0 & Ranker temp & 0.7 \\
$T$ (Chronos sampling temp) & 1.5 & $\eta$ (\texttt{prior\_temp}) & 1.0  \\
\bottomrule
\end{tabular}
\end{table}

\paragraph{Configuration guardrails.} TCS disabled on CiK; $\beta{=}1.0$ (standard uniform backup; the \texttt{WORST\_SEGMENT} attribution is dormant at $\beta{=}1$); the same continuous-Judge and Ranker prompts on both benchmarks. The cached Judge de-duplicates identical greedy prompts and memoizes verdicts across simulations; because the Judge decodes at temperature $0$, cached results are bit-identical. 

\paragraph{Trust-region pick rule (Time-MMD).}
We rank all $B$ completed rollouts by $\mathrm{value}(\tau_i)$ from Eq.~\eqref{eq:pickbest} and retain the top $k=5$. We then select $\tau^\star=\arg\min_{i\in\mathrm{top}\text{-}5}\|\tau_i-\mathrm{gmedian}(\{\tau_1,\ldots,\tau_B\})\|_2$, where the geometric median of all rollouts is computed using 25 iterations of Weiszfeld's algorithm. The geometric median represents the centre of the backbone-generated candidate pool, while the value-based shortlist restricts the choice to trajectories supported by the textual context. Selecting the shortlisted trajectory nearest this centre therefore acts as an implicit trust region, preventing a single high but noisy Judge score from selecting an unrepresentative forecast. In the w/o-MCTS ablation, where visit statistics are unavailable, $\mathrm{value}(\tau_i)$ reduces to the Judge score. Setting $k=1$ recovers direct Judge argmax selection, whereas $k=B$ selects the rollout nearest the geometric median of the full pool.

\section{\rc{} Algorithm}
\label{app:algorithm}
Algorithm~\ref{alg:recast} summarizes the complete \rc{} search procedure of \S\ref{sec:method-search}, with the default configuration of App.~\ref{app:hyperparams}.

\begin{algorithm*}[t]
\caption{\rc{}: MCTS-based planning over TSFM-generated candidates for one instance $(x,c,H)$. Default configuration: $D{=}8$, $W{=}H/D$, $K{=}8$, and $B{=}64$.}
\label{alg:recast}
\begin{algorithmic}[1]
\Require history $x$, context $c$, horizon $H$; frozen TSFM; Ranker and Judge; window size $W$, candidate pool $K$, budget $B$
\Ensure distributional or point forecast read from the search state

\State root $\gets \operatorname{node}(s_0)$, \quad
$s_0=(\varnothing,c)$, \quad
$N(\cdot,\cdot),W_{\mathrm{sum}}(\cdot,\cdot)\gets0$

\For{$b=1$ to $B$}

    \State \textit{\textbf{Selection:}} descend through expanded nodes to an expandable leaf with prefix $y_{<t}$

    \Statex \hspace{\algorithmicindent}
    decision node $s$:
    \Statex \hspace{\algorithmicindent}
    $a^\star\gets
    \arg\max_a
    \left[
    \dfrac{W_{\mathrm{sum}}(s,a)}{N(s,a)}
    +
    c_{\mathrm{puct}}\pi(a\mid s)
    \dfrac{\sqrt{\max\{1,\sum_{a'}N(s,a')\}}}{1+N(s,a)}
    \right]$

    \Statex \hspace{\algorithmicindent}
    chance node:
    $a^\star\sim\mathcal{U}\{1,\dots,K\}$

    \State \textit{\textbf{Expansion:}}
    $\{(\hat y^{(k)},\ell^{(k)})\}_{k=1}^{K}
    \gets
    \mathrm{TSFM}(x,y_{<t};K)$

    \State \textit{\textbf{Ranking:}}
    $(\texttt{relevance},\mathrm{rank}_s)
    \gets
    \textsc{Ranker}
    \bigl(c,\{\hat y^{(k)}\}_{k=1}^{K}\bigr)$

    \Statex \hspace{\algorithmicindent}
    \textbf{if} $\texttt{relevance}=\texttt{Yes}$:
    $\displaystyle
    \pi(a\mid s)\gets
    \frac{e^{-\eta\,\mathrm{rank}_s(a)}}
    {\sum_{a'}e^{-\eta\,\mathrm{rank}_s(a')}}$

    \Statex \hspace{\algorithmicindent}
    \textbf{else}: chance node,
    $a^\star\sim\mathcal{U}\{1,\dots,K\}$

    \State \textit{\textbf{Simulation:}}
    $\tau\gets
    \mathrm{TSFM}
    \bigl(x,[y_{<t};\hat y^{(a^\star)}];H\bigr)$

    \State \textit{\textbf{Backpropagation:}}
    $s(\tau)\gets
    \sum_{k=0}^{9}
    k\,P(\texttt{SCORE}{=}k\mid c,\tau)$

    \Statex \hspace{\algorithmicindent}
    $N(s,a)\mathrel{+}=1,\quad
    W_{\mathrm{sum}}(s,a)\mathrel{+}=s(\tau),
    \quad \forall (s,a)\in\mathcal{E}(\tau)$

\EndFor

\State \textbf{Output:} read a distributional or point forecast from the accumulated search state

\end{algorithmic}
\end{algorithm*}

\section{Metric Definitions}
\label{app:metrics}

\paragraph{CRPS}
For a univariate predictive distribution with CDF $F$ and an observation $z$, the continuous ranked probability score is
\[
\mathrm{CRPS}(F,z)=\int_{\mathbb{R}}\bigl(F(u)-\mathbf{1}\{u\ge z\}\bigr)^2\,du ,
\]
a strictly proper scoring rule~\cite{matheson1976crps,gneiting2007crps}: its expectation is minimized exactly when $F$ coincides with the distribution of $z$, so it rewards both calibration and sharpness. Every method on CiK is evaluated from an ensemble of $n{=}25$ trajectories. Given samples $x_1,\ldots,x_n$ from $F$, we use the unbiased ensemble estimator
\[
\widehat{\mathrm{CRPS}}(x_{1:n},z)
=\frac{1}{n}\sum_{i=1}^{n}\lvert x_i-z\rvert
-\frac{1}{2n(n-1)}\sum_{i=1}^{n}\sum_{j=1}^{n}\lvert x_i-x_j\rvert ,
\]
computed in its probability-weighted-moment form~\cite{zamo2018crps}. Multi-step forecasts are scored per timestep and averaged.

\paragraph{RCRPS}
CiK's region-of-interest CRPS~\cite{williams2024cik} is a scaled variant of the threshold-weighted CRPS~\cite{allen2023evaluating} with three components. \emph{(i) Region weighting.} When a task designates a region of interest $I\subset\{1,\dots,H\}$, the timesteps that the textual context actually constrains, the CRPS term is
$\mathrm{CRPS}_w=\tfrac{1}{2}\,\overline{\mathrm{CRPS}}_{t\in I}+\tfrac{1}{2}\,\overline{\mathrm{CRPS}}_{t\notin I}$,
where $\overline{\mathrm{CRPS}}$ denotes the mean per-timestep CRPS over the indicated set; tasks without a region of interest use the mean over all $H$ timesteps. \emph{(ii) Constraint penalty.} For tasks with an explicit numerical constraint (a maximum, minimum, or mean value), each sample trajectory $s$ receives a violation amount $v_s\ge 0$: the amount by which it breaches the constraint, averaged over the forecast timesteps and expressed in the scaled units below ($v_s{=}0$ when the constraint is satisfied). The penalty is the CRPS of the ensemble $\{10\,v_s\}_{s=1}^{n}$ against a target of zero, which is well-defined because the ground truth satisfies the constraints by construction. \emph{(iii) Scaling.} Both terms are expressed in units of the task's typical target variation through the factor $S=1/\overline{\Delta}$, where $\overline{\Delta}$ is the ground-truth range $\max(y)-\min(y)$ of the forecast window averaged over the task's instances. The final score of one instance is
\[
\mathrm{RCRPS}=S\cdot \mathrm{CRPS}_w+\widehat{\mathrm{CRPS}}\bigl(\{10\,v_s\},0\bigr),
\]
which is invariant to the overall scale of the series and comparable across tasks. Following the official evaluation, per-instance scores are capped at $5$ so that isolated catastrophic failures cannot dominate, the per-task score is the mean over the task's five instances, and the benchmark aggregate is the weighted average of per-task scores under the official task weighting (\texttt{TASK\_NAME\_TO\_WEIGHT}; the resulting per-domain weights are listed in Table~\ref{tab:cik-stats}), which prevents families of near-duplicate tasks from dominating the average. We use the official implementation throughout; lower is better.

\paragraph{Time-MMD metrics.}
For each domain $d$, a \texttt{StandardScaler} $(\mu_d,\sigma_d)$ is fit on the training split only (App.~\ref{app:datasets}), and all forecasts and ground truths are transformed as $\tilde y=(y-\mu_d)/\sigma_d$ before scoring, so that errors are comparable across domains whose raw scales differ by orders of magnitude. For a test window with horizon $H$, point forecast $\hat y$, and ground truth $y$,
\[
\mathrm{MSE}=\frac{1}{H}\sum_{h=1}^{H}(\tilde{\hat y}_h-\tilde y_h)^2 ,
\qquad
\mathrm{MAE}=\frac{1}{H}\sum_{h=1}^{H}\lvert \tilde{\hat y}_h-\tilde y_h\rvert .
\]

\paragraph{Time-MMD aggregation.} The per-domain error is the mean over the domain's evaluated windows, pooled across its four horizons; the headline aggregate is the geometric mean of the nine per-domain errors, computed per seed and averaged over the three evaluation seeds (Table~\ref{tab:tats-main}). The geometric mean is used because the per-domain errors span different magnitudes even after normalization: it responds equally to relative improvements in every domain, whereas an arithmetic mean would be dominated by the hardest domains. The nominal count is $9\times4\times50=1{,}800$ windows per seed; $1{,}793$ are evaluated, with the exclusions (the short Security pool at $H{=}12$ and the SocialGood windows with missing ground truth, App.~\ref{app:datasets}) applied identically to every method. The blend weight $\alpha$ and the shortlist width $k$ of the trust-region output rule were fixed on offline re-judged development trees before the final runs.

\section{Per-Domain Time-MMD Results}
\label{app:timemmd-perdataset}
Tables~\ref{tab:tats-agriculture} (Agriculture), \ref{tab:tats-climate} (Climate), \ref{tab:tats-economy} (Economy), \ref{tab:tats-energy} (Energy), \ref{tab:tats-environment} (Environment), \ref{tab:tats-health} (Health), \ref{tab:tats-security} (Security), \ref{tab:tats-socialgood} (SocialGood), and \ref{tab:tats-traffic} (Traffic) break the headline aggregate of Table~\ref{tab:tats-main} down by domain under the same row convention. 

\section{Budget Curves Across LLM--TSFM Pairs}
\label{app:budget-pairs}
\begin{figure*}[t]
\centering
\includegraphics[width=0.95\textwidth]{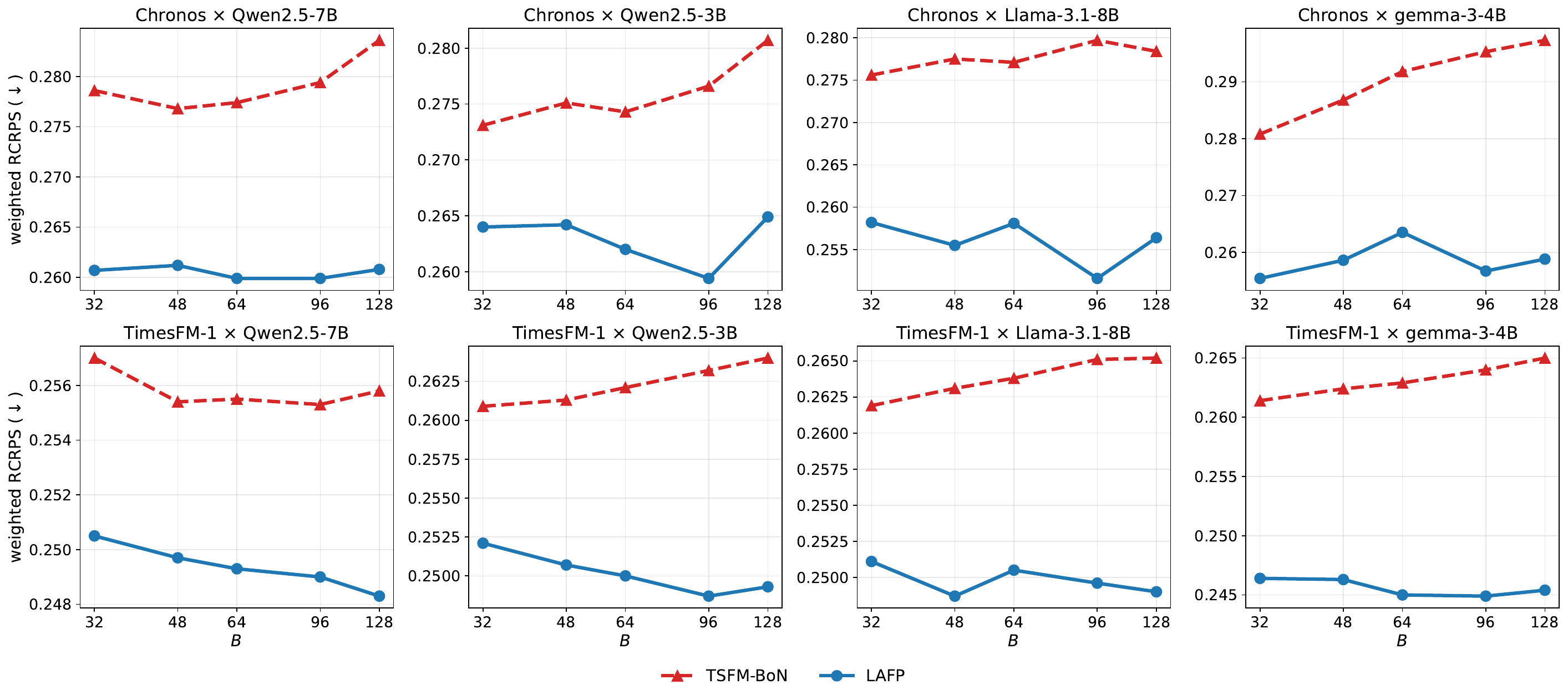}
\caption{CiK budget curves per Ranker/Judge LLM $\times$ TSFM pair: weighted RCRPS ($\downarrow$) vs.\ Judge-call budget $B$, \rc{} vs.\ TSFM-BoN at matched budget, averaged over three seeds.}
\label{fig:budget-pairs-cik}
\end{figure*}

\begin{figure*}[t]
\centering
\includegraphics[width=0.95\textwidth]{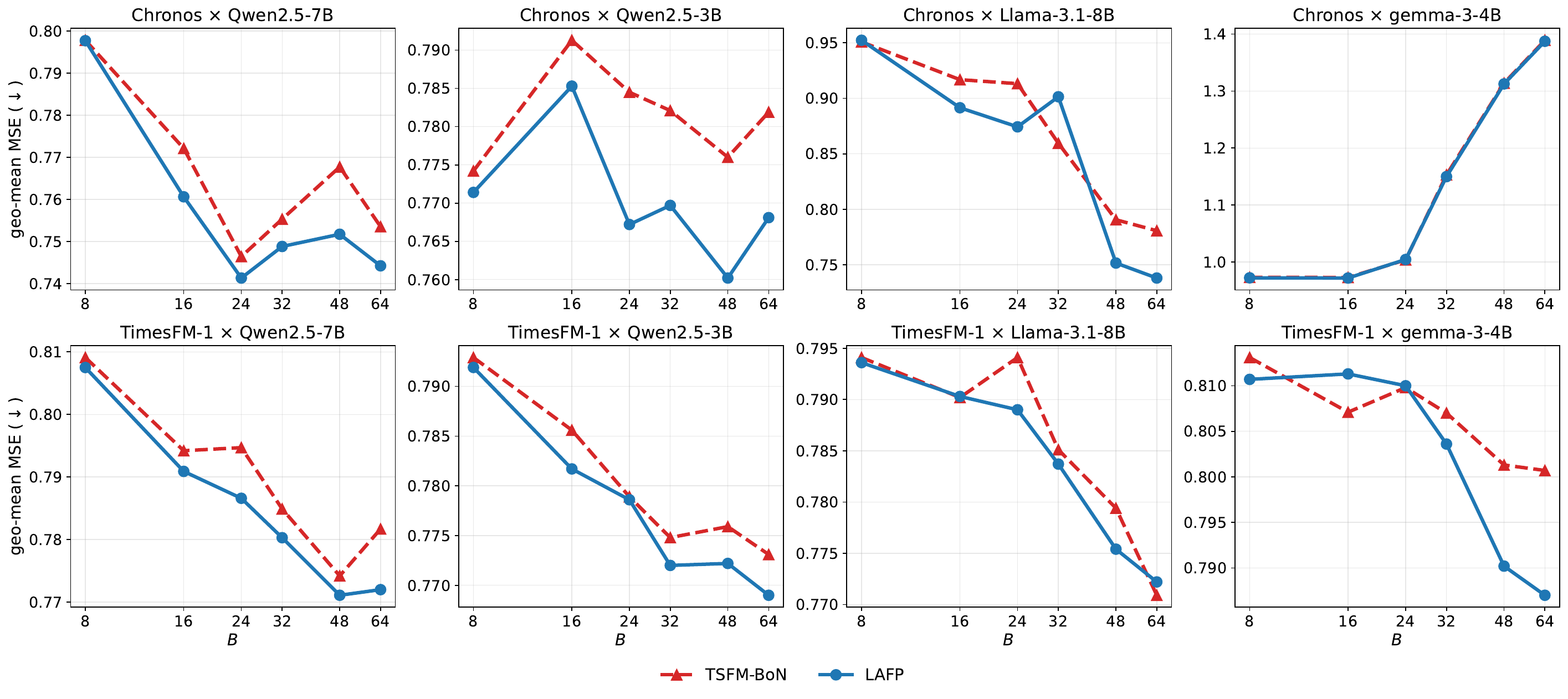}
\caption{Time-MMD budget curves per Ranker/Judge LLM $\times$ TSFM pair: geometric-mean MSE ($\downarrow$) over the nine domains vs.\ Judge-call budget $B$, \rc{} vs.\ TSFM-BoN at matched budget.}
\label{fig:budget-pairs}
\end{figure*}

Figs.~\ref{fig:budget-pairs-cik} and \ref{fig:budget-pairs} extend the budget analysis of \S\ref{sec:budget} to all eight Ranker/Judge LLM $\times$ TSFM pairs on CiK and Time-MMD, respectively. On CiK, \rc{} improves on TSFM-BoN at every evaluated budget for every pair. 

\section{LLM-as-Forecaster Divergence: Security and Health}
\label{app:timemmd-divergence}

Table~\ref{tab:domain-divergence} summarises the domain properties that predict LLM-CoT divergence. Three factors compound to produce the X cells ($\mathrm{MSE}>10^3$).

\begin{table}[h]
\centering
\caption{Time-MMD domain statistics relevant to LLM-CoT divergence. CV = coefficient of variation of the raw series (std/mean). Context = mean context length per timestep in characters. Zero-shot MSE = context-blind Chronos MSE in \texttt{StandardScaler} space. LLM-CoT MSE = Qwen2.5-7B zero-shot CoT (the strongest LLM-CoT configuration).}
\label{tab:domain-divergence}
\small
\setlength{\tabcolsep}{4pt}
\begin{tabular}{@{}lrrrr@{}}
\toprule
Domain & CV & Context (chars) & Zero-shot MSE & LLM-CoT MSE \\
\midrule
Economy    & 0.29 & 645  & 0.035  & 0.059 \\
SocialGood & 0.29 & 1244 & 0.975  & 0.720 \\
Traffic    & 0.22 & 730  & 0.294  & 0.411 \\
Agriculture & 0.13 & 442 & 34.999 & 0.915 \\
Climate    & 0.21 & 548  & 1.621  & 2.103 \\
Energy     & 0.41 & 139  & 1.804  & 33.344 \\
Environment & 0.49 & 152 & 0.476  & 4.362 \\
\midrule
Health     & 0.77 & 205  & 3.554  & 9.283 \\
Security   & 0.97 & 828  & 169.41 & 120.79 \\
\bottomrule
\end{tabular}
\end{table}

\paragraph{Factor 1: Unit-ambiguous raw scale (Security).} Security tracks food-aid beneficiary counts; raw values span 18 million to 20 billion. The context text discusses food crises and droughts but never states the numeric magnitude. An LLM reading this context plausibly infers counts in the thousands or millions, 100 to 10{,}000$\times$ smaller than the actual values. Even Qwen2.5-7B reaches MSE = 121, because any systematic scale error is squared and accumulated over all forecast steps. The weakest model diverges entirely (Qwen2.5-3B: ${\sim}6\times10^{11}$), while Llama-3.1-8B and gemma-3-4B land at 347 and 166. Compounding this, Security's test set lies $+4.0\sigma$ above the training distribution (a large surge in beneficiary counts), so even the context-blind Chronos reaches MSE = 169, the LLM-as-forecaster error stacks on top of an already broken extrapolation regime.

\paragraph{Factor 2: Short context with no numeric anchor (Health, Environment).} Health context averages only $\approx$205 characters of clinical text (e.g., ``Zanamivir has been shown effective against influenza'') with no numeric values. The target series is \% influenza-like illness (ILI), ranging 0--8\%. Without a scale anchor, Llama-3.1-8B outputs values in absolute patient-count units, reaching MSE = $3\times10^{14}$ (Qwen2.5-3B: $1\times10^{10}$). Environment (152 chars, regulatory text) shows the same failure (Llama-3.1-8B: $2.8\times10^{4}$; gemma-3-4B: $2.0\times10^{3}$). By contrast, Economy (645 chars with explicit dollar references) and SocialGood (1{,}244 chars with quantitative unemployment discussion) supply clear numeric anchors, keeping LLM-CoT MSE near or below 1.

\paragraph{Factor 3: High series volatility amplifies scale error.} Security (CV = 0.97) and Health (CV = 0.77) are the two most volatile domains. High CV means even a small fractional scale error translates to large absolute deviations, which are then squared. Low-CV domains (Agriculture CV = 0.13, Traffic CV = 0.22) remain bounded even when the LLM's inferred scale is imperfect.

\paragraph{Why \rc{} is unaffected.} The TSFM generates all numeric candidates; the LLM only evaluates which candidate is most consistent with the context linguistically. Scale ambiguity in the text degrades \emph{ranking quality} at most, not the magnitude of the output values. This is the structural advantage of the planning role over the forecaster role, and it is most visible precisely on the domains where LLM-as-forecaster diverges.

\section{Standard deviation of results across seeds}
\label{app:std}
Table~\ref{tab:std} reports the per-cell standard deviation (over the three random seeds $\{42,123,456\}$) of the main results (CiK weighted RCRPS; Time-MMD geometric-mean MSE over the nine domains). Standard deviations are computed at three seeds throughout.

\clearpage
\section{Prompts}
\label{app:prompts}
\subsection{Ranker}

\begin{lstlisting}[style=prompt]
SYSTEM: You are a forecasting assistant. A numerical time series
model has proposed {K} candidate continuations ("windows") for the
next {W} steps of a series. Every candidate is numerically plausible
according to that model. Your job is to rank ALL {K} candidates from
most to least consistent with the context, the series' recent
behaviour, and the forecast built so far.
- You never invent numbers. You only rank the given options.
- First assess RELEVANT: does the context directly affect the
  forecasted quantity? If NO, prefer options closest to the series'
  typical recent behaviour.
- If the context states an explicit constraint (e.g. a maximum,
  minimum, or fixed value), rank options that would violate it last.
- Output exactly five lines and nothing else:
  RELEVANT: YES / NO
  DIRECTION: UP / DOWN / FLAT / UNCLEAR
  CONFIDENCE: HIGH / MEDIUM / LOW
  RANKING: e.g. 3 > 1 > 4 > 2  (option numbers, each used once)
  WHY: <one sentence>
[definitions of DIRECTION and CONFIDENCE and five calibration
 examples omitted]

USER: SERIES: <description> | CONTEXT: <text, up to 2000 chars> |
RECENT HISTORY: <last 20 steps as % changes, volatility, trend> |
FORECAST SO FAR: <committed segments, or none> |
CANDIDATE WINDOWS for steps <a>-<b>: Option i = <step-wise %
changes -- pattern label> | Rank all {K} options.
\end{lstlisting}

\subsection{Judge}

\begin{lstlisting}[style=prompt]
SYSTEM: You are a forecast verifier. You see a CONTEXT and ONE
candidate FORECAST for a time series, plus the series' RECENT
HISTORY as raw values. Score the forecast 0-9.
Reason briefly in this order, then score:
1. CLAIM -- does the context state or imply anything specific and
   testable about the forecast window? Derive the expected values.
2. CLAIM CHECK -- compare the forecast's values at the affected
   steps to the target; steps outside the claim's window are never
   evidence for or against it.
3. SHAPE CHECK -- are the remaining segments a believable
   continuation of the history? With no testable claim, score by
   how closely the forecast matches the history's level, trend,
   and cycle amplitude.
Scoring (0-9): 9-8 = claimed effect clearly present with the right
direction, size, and timing; 7-6 = effect present but size or
timing somewhat off; 5-4 = mixed; 3-2 = effect absent or a clear
shape problem; 1-0 = opposite to the stated event, violates an
explicit bound, or numerically absurd. Use the FULL scale.
Output exactly two lines and nothing else:
  ANALYSIS: <2-3 short sentences following steps 1-3>
  SCORE: <single digit 0-9>

USER: <series description> | CONTEXT: <text, up to 2000 chars> |
RECENT HISTORY: <last <n> raw values; mean, volatility, trend> |
FORECAST: <H steps as segments with values; long segments
summarized by endpoints, range, and mean> | Judge the forecast.
\end{lstlisting}

\subsection{DirectPrompt}
The DirectPrompt template follows the official CiK baseline prompt~\cite{williams2024cik} and is reproduced here in full; on Time-MMD the window's paired report fills the context block. Angle brackets mark the per-instance fields.

\begin{lstlisting}[style=prompt]
SYSTEM: You are a useful forecasting assistant.

USER:
I have a time series forecasting task for you.

Here is some context about the task. Make sure to factor in any
background knowledge, satisfy any constraints, and respect any
scenarios.
<context>
<benchmark context>
</context>

Here is a historical time series in (timestamp, value) format:
<history>
(<timestamp 1>, <value 1>)
...
(<timestamp n>, <value n>)
</history>

Now please predict the value at the following timestamps:
<forecast timestamps>.

Return the forecast in (timestamp, value) format in between
<forecast> and </forecast> tags.
Do not include any other information (e.g., comments) in the
forecast.

Example: <example>

\end{lstlisting}

\clearpage
\subsection{LLM-CoT}
LLM-CoT shares DirectPrompt's template and differs only in the output instruction, which asks for explicit reasoning before the forecast; the full prompt is:

\begin{lstlisting}[style=prompt]
SYSTEM: You are a useful forecasting assistant.

USER:
I have a time series forecasting task for you.

Here is some context about the task. Make sure to factor in any
background knowledge, satisfy any constraints, and respect any
scenarios.
<context>
<benchmark context>
</context>

Here is a historical time series in (timestamp, value) format:
<history>
(<timestamp 1>, <value 1>)
...
(<timestamp n>, <value n>)
</history>

Now please predict the value at the following timestamps:
<forecast timestamps>.

Think step by step before answering.
Write your brief reasoning between <thinking> and </thinking> tags.
Then return the forecast in (timestamp, value) format in between
<forecast> and </forecast> tags.
Do not include any other information (e.g., comments) in the
forecast.

Example: <example>

\end{lstlisting}

\begin{table*}[t]
\centering
\caption{Standard deviation of the main results across three seeds. CiK weighted RCRPS $\downarrow$ (left); Time-MMD geo-mean MSE $\downarrow$ (right). }
\label{tab:std}
\scriptsize\setlength{\tabcolsep}{2.5pt}\resizebox{\textwidth}{!}{%
\begin{tabular}{@{}l cccccccc cccccccc@{}}\toprule
& \multicolumn{8}{c}{CiK} & \multicolumn{8}{c}{Time-MMD} \\
\cmidrule(lr){2-9}\cmidrule(lr){10-17}
& \multicolumn{2}{c}{Qwen2.5-3B} & \multicolumn{2}{c}{Qwen2.5-7B} & \multicolumn{2}{c}{Llama-3.1-8B} & \multicolumn{2}{c}{gemma-3-4B} & \multicolumn{2}{c}{Qwen2.5-3B} & \multicolumn{2}{c}{Qwen2.5-7B} & \multicolumn{2}{c}{Llama-3.1-8B} & \multicolumn{2}{c}{gemma-3-4B} \\
\cmidrule(lr){2-3}\cmidrule(lr){4-5}\cmidrule(lr){6-7}\cmidrule(lr){8-9}\cmidrule(lr){10-11}\cmidrule(lr){12-13}\cmidrule(lr){14-15}\cmidrule(lr){16-17}
Method & Chr. & Tim. & Chr. & Tim. & Chr. & Tim. & Chr. & Tim. & Chr. & Tim. & Chr. & Tim. & Chr. & Tim. & Chr. & Tim. \\ \midrule
Zero-shot TSFM & 0.005 & 0.004 & 0.005 & 0.004 & 0.005 & 0.004 & 0.005 & 0.004 & 0.392 & 0.040 & 0.392 & 0.040 & 0.392 & 0.040 & 0.392 & 0.040 \\
DirectPrompt & \multicolumn{2}{c}{0.010} & \multicolumn{2}{c}{0.003} & \multicolumn{2}{c}{0.022} & \multicolumn{2}{c}{0.028} & \multicolumn{2}{c}{1.501} & \multicolumn{2}{c}{0.071} & \multicolumn{2}{c}{0.831} & \multicolumn{2}{c}{0.053} \\
LLM-CoT & \multicolumn{2}{c}{0.007} & \multicolumn{2}{c}{0.005} & \multicolumn{2}{c}{0.021} & \multicolumn{2}{c}{0.025} & \multicolumn{2}{c}{2398.2} & \multicolumn{2}{c}{0.323} & \multicolumn{2}{c}{2.759} & \multicolumn{2}{c}{3.926} \\
NEXUS & \multicolumn{2}{c}{0.017} & \multicolumn{2}{c}{0.006} & \multicolumn{2}{c}{0.011} & \multicolumn{2}{c}{0.035} & \multicolumn{2}{c}{0.040} & \multicolumn{2}{c}{0.020} & \multicolumn{2}{c}{0.025} & \multicolumn{2}{c}{0.068} \\
TSFM-BoN & 0.004 & 0.004 & 0.007 & 0.003 & 0.007 & 0.003 & 0.006 & 0.003 & 0.034 & 0.033 & 0.034 & 0.033 & 0.034 & 0.033 & 0.034 & 0.033 \\
\textbf{\rc{} (ours)} & 0.006 & 0.002 & 0.006 & 0.001 & 0.010 & 0.003 & 0.014 & 0.002 & 0.067 & 0.045 & 0.029 & 0.029 & 0.037 & 0.039 & 0.313 & 0.040 \\
\bottomrule\end{tabular}}\end{table*}

\begin{table*}[t]
\centering
\caption{Time-MMD \textbf{Agriculture} (monthly) raw mean MSE ($\downarrow$, \texttt{StandardScaler} space; averaged over 3 seeds), the per-domain breakdown of Table~\ref{tab:tats-main} under the same convention. Best score per column in \textbf{bold}. The Zero-shot Chronos mean is dominated by documented deterministic wild single draws (verified bit-identical across re-runs).}
\label{tab:tats-agriculture}
\scriptsize
\setlength{\tabcolsep}{4pt}
\resizebox{0.8\textwidth}{!}{%
\begin{tabular}{@{}l cc cc cc cc@{}}
\toprule
& \multicolumn{2}{c}{Qwen2.5-3B} & \multicolumn{2}{c}{Qwen2.5-7B}
 & \multicolumn{2}{c}{Llama-3.1-8B} & \multicolumn{2}{c}{gemma-3-4B} \\
\cmidrule(lr){2-3}\cmidrule(lr){4-5}\cmidrule(lr){6-7}\cmidrule(lr){8-9}
Method & Chr. & Tim. & Chr. & Tim. & Chr. & Tim. & Chr. & Tim. \\
\midrule
\multicolumn{9}{@{}l}{\textit{TSFM only}} \\
Zero-shot TSFM & 34.999 & 0.382 & 34.999 & 0.382 & 34.999 & 0.382 & 34.999 & 0.382 \\
\midrule
\multicolumn{9}{@{}l}{\textit{LLM as forecaster}} \\
DirectPrompt & \multicolumn{2}{c}{1.224} & \multicolumn{2}{c}{0.717} & \multicolumn{2}{c}{1.253} & \multicolumn{2}{c}{0.823} \\
LLM-CoT & \multicolumn{2}{c}{4.358} & \multicolumn{2}{c}{0.915} & \multicolumn{2}{c}{470.328} & \multicolumn{2}{c}{3.115} \\
NEXUS & \multicolumn{2}{c}{0.687} & \multicolumn{2}{c}{0.355} & \multicolumn{2}{c}{\textbf{0.306}} & \multicolumn{2}{c}{\textbf{0.726}} \\
\midrule
\multicolumn{9}{@{}l}{\textit{LLM as verifier}} \\
TSFM-BoN (best $K$) & 0.290 & \textbf{0.326} & \textbf{0.306} & 0.334 & 0.271 & 0.348 & 0.959 & 0.359 \\
\textbf{\rc{} (ours)} & \textbf{0.268} & \textbf{0.326} & 0.308 & \textbf{0.327} & \textbf{0.268} & 0.348 & 0.959 & \textbf{0.336} \\
\bottomrule
\end{tabular}%
}
\end{table*}

\begin{table*}[t]
\centering
\caption{Time-MMD \textbf{Climate} (monthly) raw mean MSE ($\downarrow$, \texttt{StandardScaler} space; averaged over 3 seeds), the per-domain breakdown of Table~\ref{tab:tats-main} under the same convention. Best score per column in \textbf{bold}.}
\label{tab:tats-climate}
\scriptsize
\setlength{\tabcolsep}{4pt}
\resizebox{0.8\textwidth}{!}{%
\begin{tabular}{@{}l cc cc cc cc@{}}
\toprule
& \multicolumn{2}{c}{Qwen2.5-3B} & \multicolumn{2}{c}{Qwen2.5-7B}
 & \multicolumn{2}{c}{Llama-3.1-8B} & \multicolumn{2}{c}{gemma-3-4B} \\
\cmidrule(lr){2-3}\cmidrule(lr){4-5}\cmidrule(lr){6-7}\cmidrule(lr){8-9}
Method & Chr. & Tim. & Chr. & Tim. & Chr. & Tim. & Chr. & Tim. \\
\midrule
\multicolumn{9}{@{}l}{\textit{TSFM only}} \\
Zero-shot TSFM & 1.621 & 1.652 & 1.621 & 1.652 & 1.621 & 1.652 & 1.621 & 1.652 \\
\midrule
\multicolumn{9}{@{}l}{\textit{LLM as forecaster}} \\
DirectPrompt & \multicolumn{2}{c}{1.788} & \multicolumn{2}{c}{1.717} & \multicolumn{2}{c}{1.857} & \multicolumn{2}{c}{1.914} \\
LLM-CoT & \multicolumn{2}{c}{1.791} & \multicolumn{2}{c}{2.103} & \multicolumn{2}{c}{1.542} & \multicolumn{2}{c}{2.122} \\
NEXUS & \multicolumn{2}{c}{1.535} & \multicolumn{2}{c}{1.991} & \multicolumn{2}{c}{2.982} & \multicolumn{2}{c}{\textbf{1.153}} \\
\midrule
\multicolumn{9}{@{}l}{\textit{LLM as verifier}} \\
TSFM-BoN (best $K$) & 1.519 & \textbf{1.307} & 1.354 & \textbf{1.293} & \textbf{1.446} & \textbf{1.317} & 1.404 & 1.320 \\
\textbf{\rc{} (ours)} & \textbf{1.498} & \textbf{1.307} & \textbf{1.303} & 1.307 & \textbf{1.446} & 1.324 & 1.404 & 1.333 \\
\bottomrule
\end{tabular}%
}
\end{table*}

\begin{table*}[t]
\centering
\caption{Time-MMD \textbf{Economy} (monthly) raw mean MSE ($\downarrow$, \texttt{StandardScaler} space; averaged over 3 seeds), the per-domain breakdown of Table~\ref{tab:tats-main} under the same convention. Best score per column in \textbf{bold}.}
\label{tab:tats-economy}
\scriptsize
\setlength{\tabcolsep}{4pt}
\resizebox{0.8\textwidth}{!}{%
\begin{tabular}{@{}l cc cc cc cc@{}}
\toprule
& \multicolumn{2}{c}{Qwen2.5-3B} & \multicolumn{2}{c}{Qwen2.5-7B}
 & \multicolumn{2}{c}{Llama-3.1-8B} & \multicolumn{2}{c}{gemma-3-4B} \\
\cmidrule(lr){2-3}\cmidrule(lr){4-5}\cmidrule(lr){6-7}\cmidrule(lr){8-9}
Method & Chr. & Tim. & Chr. & Tim. & Chr. & Tim. & Chr. & Tim. \\
\midrule
\multicolumn{9}{@{}l}{\textit{TSFM only}} \\
Zero-shot TSFM & 0.035 & 0.042 & 0.035 & 0.042 & 0.035 & 0.042 & 0.035 & 0.042 \\
\midrule
\multicolumn{9}{@{}l}{\textit{LLM as forecaster}} \\
DirectPrompt & \multicolumn{2}{c}{0.048} & \multicolumn{2}{c}{0.035} & \multicolumn{2}{c}{0.059} & \multicolumn{2}{c}{0.039} \\
LLM-CoT & \multicolumn{2}{c}{0.070} & \multicolumn{2}{c}{0.059} & \multicolumn{2}{c}{0.103} & \multicolumn{2}{c}{0.105} \\
NEXUS & \multicolumn{2}{c}{0.039} & \multicolumn{2}{c}{0.056} & \multicolumn{2}{c}{0.064} & \multicolumn{2}{c}{0.038} \\
\midrule
\multicolumn{9}{@{}l}{\textit{LLM as verifier}} \\
TSFM-BoN (best $K$) & \textbf{0.028} & \textbf{0.034} & 0.029 & 0.035 & \textbf{0.028} & 0.035 & \textbf{0.029} & \textbf{0.035} \\
\textbf{\rc{} (ours)} & \textbf{0.028} & \textbf{0.034} & \textbf{0.028} & \textbf{0.034} & \textbf{0.028} & \textbf{0.034} & \textbf{0.029} & \textbf{0.035} \\
\bottomrule
\end{tabular}%
}
\end{table*}

\begin{table*}[t]
\centering
\caption{Time-MMD \textbf{Energy} (weekly) raw mean MSE ($\downarrow$, \texttt{StandardScaler} space; averaged over 3 seeds), the per-domain breakdown of Table~\ref{tab:tats-main} under the same convention. Best score per column in \textbf{bold}.}
\label{tab:tats-energy}
\scriptsize
\setlength{\tabcolsep}{4pt}
\resizebox{0.8\textwidth}{!}{%
\begin{tabular}{@{}l cc cc cc cc@{}}
\toprule
& \multicolumn{2}{c}{Qwen2.5-3B} & \multicolumn{2}{c}{Qwen2.5-7B}
 & \multicolumn{2}{c}{Llama-3.1-8B} & \multicolumn{2}{c}{gemma-3-4B} \\
\cmidrule(lr){2-3}\cmidrule(lr){4-5}\cmidrule(lr){6-7}\cmidrule(lr){8-9}
Method & Chr. & Tim. & Chr. & Tim. & Chr. & Tim. & Chr. & Tim. \\
\midrule
\multicolumn{9}{@{}l}{\textit{TSFM only}} \\
Zero-shot TSFM & 1.804 & 0.318 & 1.804 & 0.318 & 1.804 & 0.318 & 1.804 & \textbf{0.318} \\
\midrule
\multicolumn{9}{@{}l}{\textit{LLM as forecaster}} \\
DirectPrompt & \multicolumn{2}{c}{0.414} & \multicolumn{2}{c}{1.010} & \multicolumn{2}{c}{289.892} & \multicolumn{2}{c}{0.834} \\
LLM-CoT & \multicolumn{2}{c}{0.978} & \multicolumn{2}{c}{33.344} & \multicolumn{2}{c}{175.008} & \multicolumn{2}{c}{4.974} \\
NEXUS & \multicolumn{2}{c}{\textbf{0.323}} & \multicolumn{2}{c}{\textbf{0.292}} & \multicolumn{2}{c}{\textbf{0.430}} & \multicolumn{2}{c}{\textbf{0.320}} \\
\midrule
\multicolumn{9}{@{}l}{\textit{LLM as verifier}} \\
TSFM-BoN (best $K$) & 0.620 & 0.315 & 0.488 & 0.315 & 0.781 & 0.315 & 6.494 & 0.321 \\
\textbf{\rc{} (ours)} & 0.625 & \textbf{0.309} & 0.502 & 0.313 & 0.516 & \textbf{0.313} & 6.494 & 0.323 \\
\bottomrule
\end{tabular}%
}
\end{table*}

\begin{table*}[t]
\centering
\caption{Time-MMD \textbf{Environment} (daily) raw mean MSE ($\downarrow$, \texttt{StandardScaler} space; averaged over 3 seeds), the per-domain breakdown of Table~\ref{tab:tats-main} under the same convention. Best score per column in \textbf{bold}. Cells exceeding $10^3$ (divergent LLM-as-forecaster outputs) are shown as X.}
\label{tab:tats-environment}
\scriptsize
\setlength{\tabcolsep}{4pt}
\resizebox{0.8\textwidth}{!}{%
\begin{tabular}{@{}l cc cc cc cc@{}}
\toprule
& \multicolumn{2}{c}{Qwen2.5-3B} & \multicolumn{2}{c}{Qwen2.5-7B}
 & \multicolumn{2}{c}{Llama-3.1-8B} & \multicolumn{2}{c}{gemma-3-4B} \\
\cmidrule(lr){2-3}\cmidrule(lr){4-5}\cmidrule(lr){6-7}\cmidrule(lr){8-9}
Method & Chr. & Tim. & Chr. & Tim. & Chr. & Tim. & Chr. & Tim. \\
\midrule
\multicolumn{9}{@{}l}{\textit{TSFM only}} \\
Zero-shot TSFM & 0.476 & 0.415 & 0.476 & 0.415 & 0.476 & 0.415 & 0.476 & 0.415 \\
\midrule
\multicolumn{9}{@{}l}{\textit{LLM as forecaster}} \\
DirectPrompt & \multicolumn{2}{c}{4.392} & \multicolumn{2}{c}{4.801} & \multicolumn{2}{c}{1.023} & \multicolumn{2}{c}{8.143} \\
LLM-CoT & \multicolumn{2}{c}{8.082} & \multicolumn{2}{c}{4.362} & \multicolumn{2}{c}{X} & \multicolumn{2}{c}{X} \\
NEXUS & \multicolumn{2}{c}{\textbf{0.344}} & \multicolumn{2}{c}{\textbf{0.345}} & \multicolumn{2}{c}{\textbf{0.338}} & \multicolumn{2}{c}{\textbf{0.338}} \\
\midrule
\multicolumn{9}{@{}l}{\textit{LLM as verifier}} \\
TSFM-BoN (best $K$) & 0.419 & 0.382 & 0.394 & 0.390 & 0.386 & 0.373 & 0.426 & 0.389 \\
\textbf{\rc{} (ours)} & 0.409 & 0.379 & 0.392 & 0.385 & 0.385 & 0.374 & 0.426 & 0.388 \\
\bottomrule
\end{tabular}%
}
\end{table*}

\begin{table*}[t]
\centering
\caption{Time-MMD \textbf{Health} (weekly) raw mean MSE ($\downarrow$, \texttt{StandardScaler} space; averaged over 3 seeds), the per-domain breakdown of Table~\ref{tab:tats-main} under the same convention. Best score per column in \textbf{bold}. Cells exceeding $10^3$ (divergent LLM-as-forecaster outputs) are shown as X.}
\label{tab:tats-health}
\scriptsize
\setlength{\tabcolsep}{4pt}
\resizebox{0.8\textwidth}{!}{%
\begin{tabular}{@{}l cc cc cc cc@{}}
\toprule
& \multicolumn{2}{c}{Qwen2.5-3B} & \multicolumn{2}{c}{Qwen2.5-7B}
 & \multicolumn{2}{c}{Llama-3.1-8B} & \multicolumn{2}{c}{gemma-3-4B} \\
\cmidrule(lr){2-3}\cmidrule(lr){4-5}\cmidrule(lr){6-7}\cmidrule(lr){8-9}
Method & Chr. & Tim. & Chr. & Tim. & Chr. & Tim. & Chr. & Tim. \\
\midrule
\multicolumn{9}{@{}l}{\textit{TSFM only}} \\
Zero-shot TSFM & 3.554 & 3.124 & 3.554 & 3.124 & 3.554 & \textbf{3.124} & 3.554 & 3.124 \\
\midrule
\multicolumn{9}{@{}l}{\textit{LLM as forecaster}} \\
DirectPrompt & \multicolumn{2}{c}{X} & \multicolumn{2}{c}{6.132} & \multicolumn{2}{c}{142.514} & \multicolumn{2}{c}{4.284} \\
LLM-CoT & \multicolumn{2}{c}{X} & \multicolumn{2}{c}{9.283} & \multicolumn{2}{c}{X} & \multicolumn{2}{c}{287.707} \\
NEXUS & \multicolumn{2}{c}{5.358} & \multicolumn{2}{c}{\textbf{2.852}} & \multicolumn{2}{c}{4.408} & \multicolumn{2}{c}{\textbf{2.785}} \\
\midrule
\multicolumn{9}{@{}l}{\textit{LLM as verifier}} \\
TSFM-BoN (best $K$) & \textbf{2.824} & \textbf{3.109} & 2.936 & 3.112 & 2.974 & 3.143 & 5.621 & 3.075 \\
\textbf{\rc{} (ours)} & \textbf{2.824} & \textbf{3.109} & 3.012 & 3.112 & \textbf{2.775} & 3.143 & 5.621 & 3.075 \\
\bottomrule
\end{tabular}%
}
\end{table*}

\begin{table*}[t]
\centering
\caption{Time-MMD \textbf{Security} (monthly) raw mean MSE ($\downarrow$, \texttt{StandardScaler} space; averaged over 3 seeds), the per-domain breakdown of Table~\ref{tab:tats-main} under the same convention. Best score per column in \textbf{bold}. Cells exceeding $10^3$ (divergent LLM-as-forecaster outputs) are shown as X.}
\label{tab:tats-security}
\scriptsize
\setlength{\tabcolsep}{4pt}
\resizebox{0.8\textwidth}{!}{%
\begin{tabular}{@{}l cc cc cc cc@{}}
\toprule
& \multicolumn{2}{c}{Qwen2.5-3B} & \multicolumn{2}{c}{Qwen2.5-7B}
 & \multicolumn{2}{c}{Llama-3.1-8B} & \multicolumn{2}{c}{gemma-3-4B} \\
\cmidrule(lr){2-3}\cmidrule(lr){4-5}\cmidrule(lr){6-7}\cmidrule(lr){8-9}
Method & Chr. & Tim. & Chr. & Tim. & Chr. & Tim. & Chr. & Tim. \\
\midrule
\multicolumn{9}{@{}l}{\textit{TSFM only}} \\
Zero-shot TSFM & 169.407 & 75.376 & 169.407 & 75.376 & 169.407 & 75.376 & 169.407 & 75.376 \\
\midrule
\multicolumn{9}{@{}l}{\textit{LLM as forecaster}} \\
DirectPrompt & \multicolumn{2}{c}{163.218} & \multicolumn{2}{c}{128.036} & \multicolumn{2}{c}{121.772} & \multicolumn{2}{c}{111.304} \\
LLM-CoT & \multicolumn{2}{c}{X} & \multicolumn{2}{c}{120.790} & \multicolumn{2}{c}{347.472} & \multicolumn{2}{c}{166.429} \\
NEXUS & \multicolumn{2}{c}{\textbf{72.025}} & \multicolumn{2}{c}{78.223} & \multicolumn{2}{c}{76.991} & \multicolumn{2}{c}{\textbf{70.971}} \\
\midrule
\multicolumn{9}{@{}l}{\textit{LLM as verifier}} \\
TSFM-BoN (best $K$) & 80.566 & \textbf{71.673} & \textbf{74.189} & 71.386 & 73.990 & 71.622 & 127.436 & 71.221 \\
\textbf{\rc{} (ours)} & 77.763 & 71.802 & \textbf{74.189} & \textbf{71.275} & \textbf{73.703} & \textbf{71.549} & 127.436 & 71.328 \\
\bottomrule
\end{tabular}%
}
\end{table*}

\begin{table*}[t]
\centering
\caption{Time-MMD \textbf{SocialGood} (monthly) raw mean MSE ($\downarrow$, \texttt{StandardScaler} space; averaged over 3 seeds), the per-domain breakdown of Table~\ref{tab:tats-main} under the same convention. Best score per column in \textbf{bold}.}
\label{tab:tats-socialgood}
\scriptsize
\setlength{\tabcolsep}{4pt}
\resizebox{0.8\textwidth}{!}{%
\begin{tabular}{@{}l cc cc cc cc@{}}
\toprule
& \multicolumn{2}{c}{Qwen2.5-3B} & \multicolumn{2}{c}{Qwen2.5-7B}
 & \multicolumn{2}{c}{Llama-3.1-8B} & \multicolumn{2}{c}{gemma-3-4B} \\
\cmidrule(lr){2-3}\cmidrule(lr){4-5}\cmidrule(lr){6-7}\cmidrule(lr){8-9}
Method & Chr. & Tim. & Chr. & Tim. & Chr. & Tim. & Chr. & Tim. \\
\midrule
\multicolumn{9}{@{}l}{\textit{TSFM only}} \\
Zero-shot TSFM & 0.975 & 1.221 & 0.975 & 1.221 & 0.975 & 1.221 & 0.975 & 1.221 \\
\midrule
\multicolumn{9}{@{}l}{\textit{LLM as forecaster}} \\
DirectPrompt & \multicolumn{2}{c}{2.422} & \multicolumn{2}{c}{1.052} & \multicolumn{2}{c}{2.423} & \multicolumn{2}{c}{0.803} \\
LLM-CoT & \multicolumn{2}{c}{\textbf{0.730}} & \multicolumn{2}{c}{\textbf{0.720}} & \multicolumn{2}{c}{\textbf{0.765}} & \multicolumn{2}{c}{\textbf{0.708}} \\
NEXUS & \multicolumn{2}{c}{1.178} & \multicolumn{2}{c}{1.492} & \multicolumn{2}{c}{1.494} & \multicolumn{2}{c}{1.042} \\
\midrule
\multicolumn{9}{@{}l}{\textit{LLM as verifier}} \\
TSFM-BoN (best $K$) & 0.738 & 1.098 & 0.814 & 1.141 & \textbf{0.693} & 1.066 & 0.788 & 1.056 \\
\textbf{\rc{} (ours)} & \textbf{0.729} & 1.083 & 0.761 & 1.132 & \textbf{0.693} & 1.088 & 0.788 & 1.056 \\
\bottomrule
\end{tabular}%
}
\end{table*}

\begin{table*}[t]
\centering
\caption{Time-MMD \textbf{Traffic} (monthly) raw mean MSE ($\downarrow$, \texttt{StandardScaler} space; averaged over 3 seeds), the per-domain breakdown of Table~\ref{tab:tats-main} under the same convention. Best score per column in \textbf{bold}.}
\label{tab:tats-traffic}
\scriptsize
\setlength{\tabcolsep}{4pt}
\resizebox{0.8\textwidth}{!}{%
\begin{tabular}{@{}l cc cc cc cc@{}}
\toprule
& \multicolumn{2}{c}{Qwen2.5-3B} & \multicolumn{2}{c}{Qwen2.5-7B}
 & \multicolumn{2}{c}{Llama-3.1-8B} & \multicolumn{2}{c}{gemma-3-4B} \\
\cmidrule(lr){2-3}\cmidrule(lr){4-5}\cmidrule(lr){6-7}\cmidrule(lr){8-9}
Method & Chr. & Tim. & Chr. & Tim. & Chr. & Tim. & Chr. & Tim. \\
\midrule
\multicolumn{9}{@{}l}{\textit{TSFM only}} \\
Zero-shot TSFM & 0.294 & 0.266 & 0.294 & 0.266 & 0.294 & 0.266 & 0.294 & 0.266 \\
\midrule
\multicolumn{9}{@{}l}{\textit{LLM as forecaster}} \\
DirectPrompt & \multicolumn{2}{c}{0.401} & \multicolumn{2}{c}{0.238} & \multicolumn{2}{c}{0.284} & \multicolumn{2}{c}{\textbf{0.187}} \\
LLM-CoT & \multicolumn{2}{c}{2.154} & \multicolumn{2}{c}{0.411} & \multicolumn{2}{c}{0.350} & \multicolumn{2}{c}{2.390} \\
NEXUS & \multicolumn{2}{c}{0.241} & \multicolumn{2}{c}{0.391} & \multicolumn{2}{c}{0.562} & \multicolumn{2}{c}{0.238} \\
\midrule
\multicolumn{9}{@{}l}{\textit{LLM as verifier}} \\
TSFM-BoN (best $K$) & \textbf{0.204} & 0.231 & 0.194 & 0.231 & \textbf{0.215} & \textbf{0.215} & 0.311 & 0.288 \\
\textbf{\rc{} (ours)} & \textbf{0.204} & \textbf{0.229} & \textbf{0.189} & \textbf{0.220} & \textbf{0.215} & \textbf{0.215} & 0.311 & 0.259 \\
\bottomrule
\end{tabular}%
}
\end{table*}

\end{document}